\documentclass[10pt,twocolumn,letterpaper]{article}

\usepackage{iccv}
\usepackage{times}
\usepackage{epsfig}
\usepackage{graphicx}
\usepackage{amsmath}
\usepackage{amssymb}
\usepackage{booktabs}
\usepackage{bm}
\usepackage{bbm}
\usepackage{setspace}
\usepackage{caption}
\usepackage{multirow}
\usepackage{extarrows}
\usepackage[normalem]{ulem}

\graphicspath{{./figures/}}

\iccvfinalcopy

\makeatletter
\let\MYcaption\@makecaption
\makeatother

\usepackage{subcaption}

\makeatletter
\let\@makecaption\MYcaption
\makeatother

\usepackage[pagebackref=true,breaklinks=true,letterpaper=true,colorlinks,bookmarks=false]{hyperref}

\usepackage[capitalize]{cleveref}
\crefname{section}{Sec.}{Secs.}
\Crefname{section}{Section}{Sections}
\Crefname{table}{Table}{Tables}
\crefname{table}{Tab.}{Tabs.}

\DeclareMathOperator*{\argmin}{arg\,min}

\newcommand{\z}{\bm{z}}
\newcommand{\zpls}{\bm{z}^{+}}
\newcommand{\zmpls}{\bm{z}_{M+}}
\newcommand{\fz}{\bm{z}^{\ast}}
\newcommand{\f}{\bm{f}}
\newcommand{\w}{\bm{w}}
\newcommand{\wpls}{\bm{w}^{+}}
\newcommand{\wmpls}{\bm{w}_{M+}}
\newcommand{\fw}{\bm{w}^{\ast}}
\newcommand{\PS}{\mathcal{P}}
\newcommand{\PNS}{\mathcal{P}_{\mathcal{N}}}
\newcommand{\PPNS}{\mathcal{P}^{+}_{\mathcal{N}}}
\newcommand{\FWS}{\mathcal{F}/\mathcal{W}^{+}}
\newcommand{\WS}{\mathcal{W}}
\newcommand{\WPS}{\mathcal{W}^{+}}
\newcommand{\ZS}{\mathcal{Z}}
\newcommand{\ZPS}{\mathcal{Z}^{+}}
\newcommand{\FZS}{\mathcal{F}/\mathcal{Z}^{+}}
\newcommand{\FSS}{\mathcal{F}/\mathcal{S}}
\newcommand{\SSp}{\mathcal{S}}
\newcommand{\FS}{\mathcal{F}}
\newcommand{\FZ}{\mathcal{F}/\mathcal{Z}}
\newcommand{\X}{\mathcal{X}}
\newcommand{\x}{\bm{x}}
\newcommand{\Lrecon}{L_\textrm{recon}}
\newcommand{\Lmse}{L_\textrm{MSE}}
\newcommand{\Lper}{L_\textrm{per}}
\newcommand{\Lenc}{L_\textrm{enc}}
\newcommand{\Lreg}{L_\textrm{reg}}

\def\iccvPaperID{11551} %

\begin{document}

\title{Revisiting Latent Space of GAN Inversion for Real Image Editing}

\author{Kai Katsumata\\
The University of Tokyo\\  
{\tt\small katsumata@nlab.ci.i.u-tokyo.ac.jp}
\and
Duc Minh Vo\\
The University of Tokyo\\
{\tt\small vo@nlab.ci.i.u-tokyo.ac.jp}
\and
Bei Liu\\
Microsoft Research\\
{\tt\small bei.liu@microsoft.com}
\and
Hideki Nakayama\\
The University of Tokyo\\
{\tt\small nakayama@nlab.ci.i.u-tokyo.ac.jp}}

\maketitle


\begin{abstract}
The exploration of the latent space in StyleGANs and GAN inversion exemplify impressive real-world image editing, yet the trade-off between reconstruction quality and editing quality remains an open problem.
In this study, we revisit StyleGANs' hyperspherical prior $\ZS$ and combine it with highly capable latent spaces to build combined spaces that faithfully invert real images while maintaining the quality of edited images.
More specifically, we propose $\FZS$ space consisting of two subspaces: $\FS$ space of an intermediate feature map of StyleGANs enabling faithful reconstruction
and $\ZPS$ space of an extended StyleGAN prior supporting high editing quality.
We project the real images into the proposed space to obtain the inverted codes, by which we then move along $\ZPS$, enabling semantic editing without sacrificing image quality.
Comprehensive experiments 
show that $\ZPS$ can replace the most commonly-used $\WS$, $\WPS$, and $\SSp$ spaces while preserving reconstruction quality, 
resulting in reduced distortion of edited images.

\end{abstract}

\section{Introduction}
\label{sec:intro}
    
The combination of GAN inversion~\cite{abdal2019image2stylegan,abdal2020image2stylegan++,Kang_2021_ICCV,feng2022near,roich2021pivotal,xia2022gan,parmar2022spatially,zhu2021barbershop,bermano2022state,zhu2020domain} and latent space editing~\cite{Shen_2020_CVPR,NEURIPS2020_ganspace,Shen_2021_CVPR} enables us to edit a wide range of real image attributes such as aging, expression, and light condition, by applying editing operations in the latent space of GANs~\cite{Shen_2020_CVPR,NEURIPS2020_ganspace,Shen_2021_CVPR} to inverted latent codes.
To this end, many methods
~\cite{abdal2019image2stylegan,abdal2020image2stylegan++,Kang_2021_ICCV,feng2022near} aiming to find the latent code of StyleGANs~\cite{Karras2019style,Karras2020analyzing,Karras2020training,Karras2021alias} that generates a given image have been developed. Recent efforts~\cite{tov2021designing,Kang_2021_ICCV,feng2022near,wu2021stylespace} have focused on reducing reconstruction loss by exploring novel embedding spaces, encoding methods, and optimization algorithms.

Nevertheless, it is still challenging to trade-off between reconstruction quality and editing quality of the edited images through GAN inversion. 
Using popular embedding spaces such as $\WS$~\cite{Karras2020analyzing}, $\WPS$~\cite{abdal2019image2stylegan}, and $\SSp$~\cite{wu2021stylespace} is able to improve the reconstruction quality, yet the low-quality edited image is unavoidable.
Recent attempts (\eg, SAM~\cite{parmar2022spatially}, PTI~\cite{roich2021pivotal}, and $\PS$~\cite{zhu2020improved}) aim to maintain the perceptual quality during semantic editing.
However, since they use transformed spaces such as $\WS$ or $\WPS$, namely unbounded embedding space with unknown boundaries, the shape of such space is too complex to edit.
Such an unbounded embedding space cannot guarantee that the edited embedding is always present in the embedding space, resulting in distortion after editing.
Unlike typically used spaces, StyleGAN's prior space $\ZS$ is a bounded embedding space, meaning that it is rich in editing quality while poor in reconstruction quality.
To address the trade-off challenge, one solution is to flexibly use the rich embedding space and the original latent space, which is the main target of this work.

To begin with, we revisit the original latent space $\ZS$, which can be easily editable. Since the latent code $\z \in \ZS$ is sampled from the hypersphere, 
the latent code can move on $\ZS$ with closed operations. 
To maintain the reconstruction quality while leveraging the robust nature of $\ZS$, we first extend $\ZS$ to $\ZPS$ and combine $\ZPS$ with a feature space $\FS$ in the output of an intermediate layer in the StyleGAN generator, proposing an alternative space namely $\FZS$.
Our proposed $\FZS$ space achieves excellent reconstruction performance due to the use of the feature space $\FS$ and increases editing quality with the aid of the hyperspherical prior of the original latent space $\ZPS$, simultaneously. Here, editing quality denotes the perceptual quality of images after performing editing operations in the latent space.

Qualitative and quantitative evaluations show that our proposed method maintains image quality after performing editing operations without sacrificing reconstruction quality.
Our main contributions are as follows:  
\begin{itemize}
  \item We revisit $\ZS$ space for GAN inversion and combine the extended revisited space $\ZPS$ and highly capable latent space to guarantee high editing quality while maintaining reconstruction quality for real image editing via GAN inversion, presenting a novel $\FZS$ space.
\item  According to the experimental results, our $\FZS$ space achieves competitive reconstruction quality with existing spaces and outperforms baselines on editing quality.
  \item We extend the method by leveraging $\ZPS$ on cutting-edge GAN inversion approaches, demonstrating improvements in editing quality over baselines.
\end{itemize}

\section{Related work}
\label{sec:relatedwork}

\noindent \textbf{GAN inversion} aims to project real images into low-dimensional latent codes, which can be mainly classified into two types of approaches: encoder-based and optimization-based approaches.
The former type of approach~\cite{tov2021designing,richardson2021encoding} trains an encoder network with a loss function that measures the distance between the input image and the reconstructed image obtained from the predicted latent code. 
The latter one~\cite{abdal2019image2stylegan,abdal2020image2stylegan++,roich2021pivotal} directly optimizes the latent code to reconstruct a target input.
A hybrid approach~\cite{Kang_2021_ICCV,parmar2022spatially}, which combines encoder-based and optimization-based approaches, initializes a latent code with the encoder prediction and then fine-tunes the latent code using an optimization method.

Pioneering studies of GAN inversion aim at the faithful reconstruction of target images.
For example, extending a latent space leads to high reconstruction quality~\cite{abdal2019image2stylegan,abdal2020image2stylegan++,Kang_2021_ICCV}.
Recently, Kang \etal\cite{Kang_2021_ICCV} and Feng \etal\cite{feng2022near} improve the robustness to out-of-range images. 
Another recent direction of GAN inversion is to increase robustness to downstream tasks~\cite{zhu2020improved}.
Zhu \etal\cite{zhu2020improved} aims to increase the editing quality of found latent codes by the regularization that directs latent codes toward a high-density region.
Although our study has the same goal as \cite{zhu2020improved}, our approach is to use a latent space where we know the bound of the codes, unlike~\cite{zhu2020improved}.

\noindent \textbf{Semantic image editing}~\cite{NEURIPS2020_ganspace,Shen_2020_CVPR,choi2022not,zhu2021low,Shen_2021_CVPR,li2021surrogate} is one of the downstream tasks of embedding real images into latent space.
The task modifies a latent code along semantically meaningful directions to generate an intended image. Supervised~\cite{Shen_2020_CVPR} and unsupervised~\cite{NEURIPS2020_ganspace,Shen_2021_CVPR} approaches are investigated to explore semantic directions. GANSpace~\cite{NEURIPS2020_ganspace} finds useful directions by computing eigenvectors on the empirical distribution of latent code. SeFa~\cite{Shen_2021_CVPR} factorizes the weights of layers that feed on latent codes. The above methods explore global semantic directions, which are shared among the latent codes and are called global methods. Another research direction is local methods~\cite{choi2022not,zhu2021low}. The local methods explore semantic directions with respect to each latent code.

\section{Approach}\label{sec:method}

In this section, we first review various latent spaces for GAN inversion, $\ZS$, $\ZPS$, $\WS$, $\WPS$, $\SSp$, $\FWS$, $\FSS$, and $\PPNS$, and their pros and cons.
Then, we introduce our proposed latent space $\FZS$ which aims to improve editing quality while maintaining reconstruction quality.

\subsection{Analysis of StyleGAN Spaces}\label{sec:fzspace}

 \begin{figure*}[htb]
   \centering
   \begin{minipage}{0.29\linewidth}
      \includegraphics[height=19em]{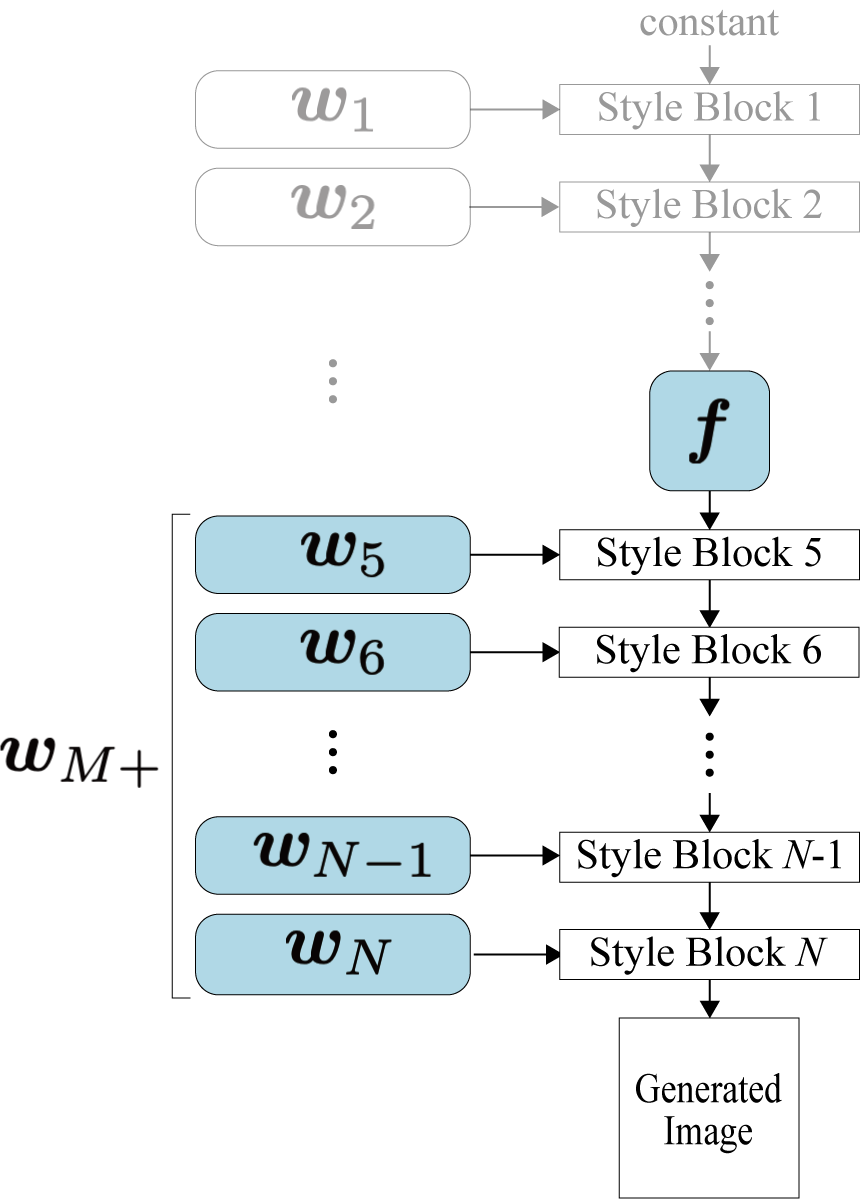}
      \subcaption{$\mathcal{F}/\mathcal{W}^{+}$ space~\cite{Kang_2021_ICCV} \label{fig:fwpspace}}
   \end{minipage}%
   \hfill
   \begin{minipage}{0.23\linewidth}
      \includegraphics[height=19em]{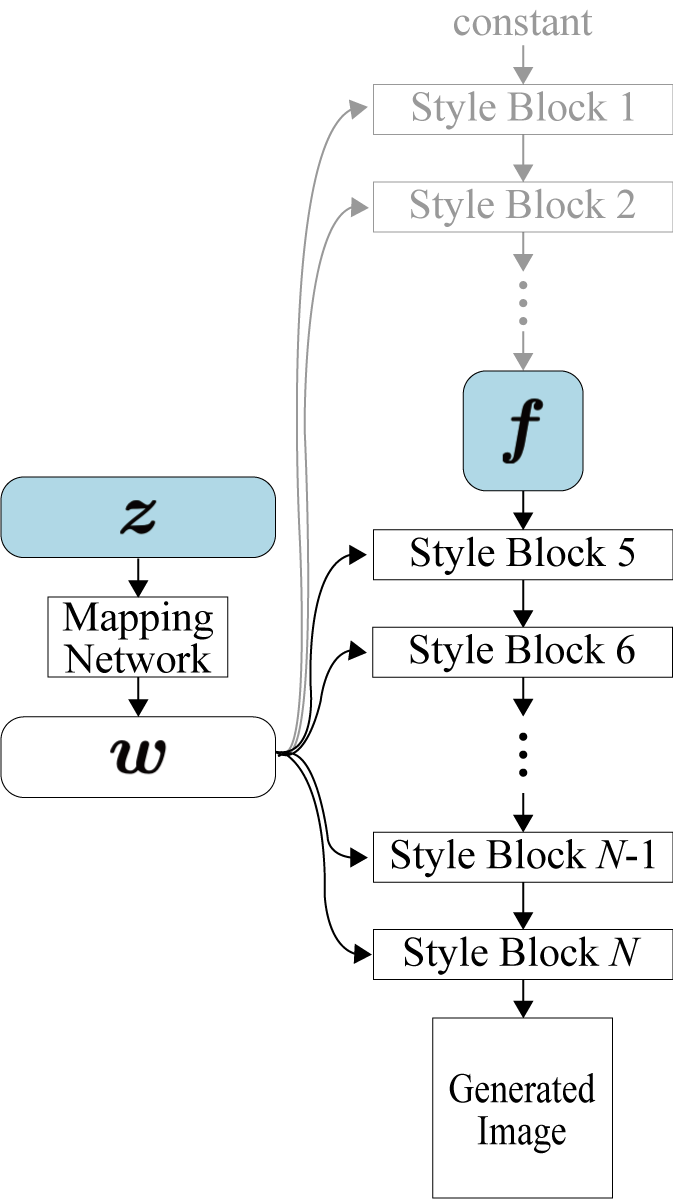}
      \subcaption{$\mathcal{F}/\mathcal{Z}$ space (ours)\label{fig:fzspace}}
   \end{minipage}%
   \hfill
   \begin{minipage}{0.48\linewidth}
      \includegraphics[height=19em]{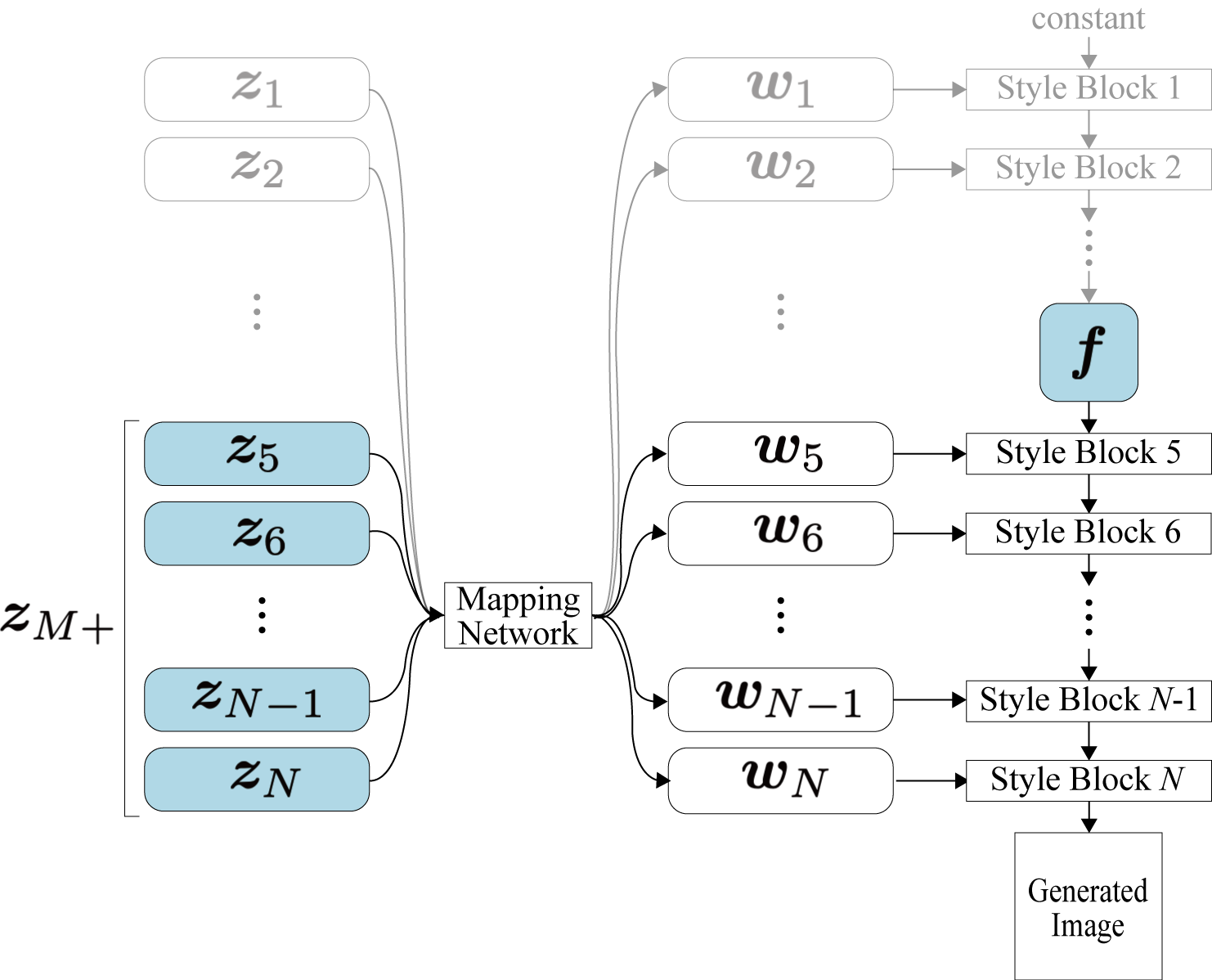}
      \subcaption{$\mathcal{F}/\mathcal{Z}^{+}$ space (ours)\label{fig:fzpspace}}
   \end{minipage}%
    \caption{Latent spaces of StyleGANs. The space $\FWS$ introduced by Kang \etal\cite{Kang_2021_ICCV} consists of $\FS$ and $\WPS$ spaces, and it leads to the faithful reconstruction of even out-of-range images. Using $\ZS$ or $\ZPS$ instead of $\WPS$, we propose $\FZ$ and $\FZS$ spaces without sacrificing reconstruction quality with the aid of $\FS$.
    The base code $\f$ is an intermediate output of the StyleGAN generator with spatial dimensions, and 
    the detail code $\wmpls$ or $\zmpls$ is a subset of
    $\wpls$ or $\zpls$ and the inputs of the upper stages of the generator. The optimizing codes are highlighted in blue.}
\end{figure*}

\noindent \textbf{$\ZS$ and $\ZPS$ Space}.
The generator $G: \ZS \to \X$ learns to map a simple distribution, called latent space $\ZS$, to the image space, 
where $\x \in \X$ is an image, and $\z \in \ZS$ is uniformly sampled from a hypersphere.
The primitive latent code of the StyleGAN family is with $512$ dimensions.

AgileGAN~\cite{song2021agilegan} and StyleAlign~\cite{wu2021stylealign} employ the extended space $\ZPS$, which provides a different latent code from $\ZS$ for each layer.
Unlike GAN inversion, AgileGAN~\cite{song2021agilegan} addresses stylizing portraits and does not aim at faithful reconstruction. StyleAlign~\cite{wu2021stylealign} demonstrates $\ZPS$ cannot achieve faithful reconstruction for real image inversion.
As shown in \cite{song2021agilegan,wu2021stylealign}, 
while the $\ZS$ and $\ZPS$ spaces have the issue of insufficient reconstruction quality,
the bounded nature of the spaces (\ie, hypersphere) leads to edited images with less deterioration of quality.

\noindent \textbf{$\WS$, $\WPS$, and $\SSp$ Space}. 
StyleGANs also use the intermediate latent space $\WS$ where each $\w \in \WS$ is produced by a mapping network consisting of eight fully-connected layers denoted as $\mathcal{M}: \ZS \to \WS$.
Thanks to multiple affine transformations and non-linearity functions in the mapping network $\mathcal{M}$, the features of $\WS$ are more disentangled than those of $\ZS$.

Thereafter, \cite{abdal2019image2stylegan,abdal2020image2stylegan++} introduced $\WPS$ space, achieving lower reconstruction loss by allowing to control of details of images.
Each element $\wpls$ in $\WPS$ is defined as $\wpls = \{\w_1,\w_2,\ldots,\w_N\}$, where $\w_i\in\WS$, and $N$ is the number of layers in generator that takes $\w$
as input.
For a $1024\times 1024$ StyleGAN, we use 18 different 512-dimensional latent codes as $\wpls$.
$\SSp$ space~\cite{wu2021stylespace} is spanned by style parameters, which is transformed from $\w \in \WS$ using different learned affine transformations for each layer of the generator.

Although the $\WS$, $\WPS$, and $\SSp$ spaces derive faithful reconstruction quality, distortions and artifacts may appear in edited images~\cite{wulff2020improving,zhu2020improved,song2021agilegan}.
This is because the embeddings with these spaces for the images may not correspond appropriately to $\ZPS$, the StyleGAN prior latent space, 
and the space cannot guarantee that the edited latent code reaches the original space. In this study, we aim at latent editing without suffering quality deterioration with the aid of the nature of $\ZS$ or $\ZPS$.

\noindent $\PPNS$ \textbf{Space}. 
Zhu \etal\cite{zhu2020improved} introduced a normalized space $\PS$ to explore the GAN inversion trade-offs.
The space is the deactivated space of $\WS$, which is computed by using the inversion of Leaky ReLU.
Zhu \etal\cite{zhu2020improved} also proposed the $\PNS$ space by whitening the $\PS$ space with the principal component analysis (PCA) parameter computed on a million latent codes in the $\PS$ space.
Since the distribution of $\PNS$ can be interpreted as the Gaussian distribution with zero mean and unit variance, we can calculate the density of the latent codes in $\PNS$.
As discussed in \cite{menon2020pulse}, the regularization on $\PNS$ space ensures realistic GAN inversion outcomes.
Since $\PNS$ space cannot achieve sufficient reconstruction quality, the $\PNS$ can be extended to $\PPNS$ in the same manner as $\WPS$~\cite{abdal2019image2stylegan,abdal2020image2stylegan++}.

Although $\PPNS$ improves the robustness of the reconstructed latent codes, editing operations are performed on $\WS$ or $\WPS$ space.
This means that utilizing the regularization on $\PNS$ leads to maintaining the image quality only within the neighborhood of an inverted code.
Hence, the weaknesses of the unbounded spaces as discussed above remain.
We thus seek the latent space that also guarantees generation quality in the editing phase.

\noindent $\FWS$ \textbf{and} $\FSS$ \textbf{Space}. Kang \etal\cite{Kang_2021_ICCV} proposed $\FWS$ space for generalization performance to out-of-range images (\cref{fig:fwpspace}), consisting of the $\FS$ and $\WPS$ spaces, and the space also investigated in SAM~\cite{parmar2022spatially} and Barbershop~\cite{zhu2021barbershop}. 
The coarse-scale feature map $\f \in \FS$ is an intermediate output of a generator before taking fine-scale latent codes $\wmpls = \{\w_M,\w_{M+1}\ldots,\w_N\}$.
An element $\fw = (\f, \wmpls)$ of $\FWS$ consists of a base code $\f$ and a detail code $\wmpls$. 
The information of a noise input and bottom latent codes $\{\w_1,\w_2,\ldots,\w_{M-1}\}$ is contained in $\f$, and the feature map controls the geometric information.
Kang \etal\cite{Kang_2021_ICCV} have integrated the regularization on the $\PNS$ space into the $\FWS$ space to obtain robust latent codes. The combined space has extended the range of images that can be inverted. 
$\FSS$ space~\cite{yao2022style} employs $\SSp$ space~\cite{wu2021stylespace} as an alternative to $\WPS$.

Whereas these spaces achieve faithful reconstruction quality, it has the same limitations as $\WS$, $\WPS$, $\SSp$, and $\PPNS$. 
This is because latent editing is performed on unbounded spaces. %
Thus, to continue using $\ZS$ space, we leverage the $\FS$ space, which complements the lack of representative capacity of the bounded latent space (\cref{fig:fzspace,fig:fzpspace}).

\subsection{\texorpdfstring{$\FZS$}{F/Z+} Space}

Overall, there is still no existing latent space that can guarantee both reconstruction quality and editing quality. As discussed in \cite{menon2020pulse,zhu2020improved,song2021agilegan,wulff2020improving}, leveraging $\ZS$ or $\ZPS$ spaces lead to high editing quality in exchange for reconstruction quality.
To overcome their limitation, we propose an alternative latent space $\FZS$ by extending the StyleGAN prior $\ZS$.
The proposed space consists of the feature space $\FS$ for increasing representative capacity and a bounded latent space $\ZPS$ for maintaining editing quality. We use the combination of the spaces $\FS$ and $\ZPS$
because we cannot achieve sufficient reconstruction quality when using the space $\ZPS$ to increase the editing quality.

The latent space $\FZS$ can do semantic editing safety while maintaining the reconstruction quality (\cref{fig:fzpspace}).
The $\ZPS$ space is a concatenation of the $N$ latent code $\z$, and it is derived in the same fashion as $\WPS$. 
Each element $\zpls$ in $\ZPS$ is defined as $\zpls = \{\z_1,\z_2,\ldots,\z_N\}$ where $\z_i\in\ZS$. A code $\z_i$ is an input for each layer of a StyleGAN generator and is transformed by the mapping network and AdaIN~\cite{huang2017arbitrary} before being fed into the generator.
The number of layers is $N=18$ for a $1024\times 1024$ StyleGAN. We define $\FZS$ by combining the $\FS$ and $\ZPS$ spaces.
Each element $\fz \in \FZS$ is defined as $\fz = (\f, \zmpls)$, where $\zmpls = \{\z_M,\z_{M+1},\ldots,\z_N\}$ is a set of latent codes for the fine scales of the generator. We also introduce the $\FZ$ space for comparison, consisting of $\FS$ and $\ZS$ instead of $\ZPS$.

The $\FZS$ space has the desirable properties required for GAN inversion: high reconstruction quality and high editing quality.
Our proposed space is equipped with high reconstruction capability attributed to the feature space $\FS$ and high editing quality attributed to the primitive space $\ZS$.
PULSE~\cite{menon2020pulse} discussed the importance of
considering a manifold of a latent space, which controls the content quality. Following this discussion, Zhu \etal\cite{zhu2020improved} assumes that
the deactivated $\WS$ follows a Gaussian distribution and picks a latent code located in a high-density region.
To greatly benefit from considering the latent manifold, we employ bounded latent codes. Since we know the shape of $\ZS$,
we completely utilize the information of the distribution of $\ZS$.

\subsection{Inversion to \texorpdfstring{$\FZS$}{F/Z+}}
We introduce an inversion algorithm that projects images to the $\FZS$ space.
Our method is a hybrid method that first projects a target image to $\FS$ by using a pre-trained encoder to obtain an initial 
base code and then directly optimizes the base and detail codes.

Given an input image $\x$, we find a latent code $\fz$ that reconstructs $\x$ by optimizing an objective function $\Lrecon$, which is
defined as:
\begin{align}
\Lrecon(\fz) = \Lmse(\fz) + \lambda_{\textrm{per}}\Lper(\fz),
\end{align}
where $\Lmse$ and $\Lper$ are mean squared error (MSE) and perceptual losses~\cite{ledig2017photo,zhang2018unreasonable}, respectively. The hyperparameter $\lambda_{\textrm{per}}$ controls the contributions
of $\Lmse$ and $\Lper$.
MSE loss is defined as $\Lmse(\fz) = \|\x - G(\fz) \|^{2}$, and perceptual loss is defined as $\Lper(\fz) = \|\phi(\x) - \phi(G(\fz))\|^{2}$ with an LPIPS network $\phi$,
which is a pre-training network with the VGG backbone and extracts high-level feature representation.
MSE loss and perceptual loss are the distances between the target and inverted images in the data space and feature space, respectively. We use perceptual loss to enhance 
reconstruction quality and specially to avoid blurred images~\cite{park2019semantic,ledig2017photo,zhang2018unreasonable}.

For efficient optimization, we initialize the base code by employing an encoder. We first compute a rough base code using an encoder and then optimize
a precise base code. We train an encoder $E\!:\!\X\!\to\!\FS$ by optimizing the loss function:
\begin{align}
\Lenc  =& \|G(E(\x_{\downarrow}), \bm{w}^{s}_{M+})- \x\|^{2} \\\nonumber
& + \lambda_{\textrm{enc}} \|F(G(E(\x_{\downarrow})), \bm{w}^{s}_{M+}) - F(\x)\|^{2},
\end{align}
where $G: \f \times \wmpls \mapsto \x$ is a generator (we use a different notation than the above-mentioned $G$ in Section 3 to emphasize the various inputs), and $\f^{s}$ and $\bm{w}^{s}_{M+}$ are sampled codes corresponding to sampled latent code $\z \in \ZS$.
Training images $\x = G(\f^{s}, \bm{w}^{s}_{M+})$ are reconstructed from the sampled codes $\f^{s}$ and $\bm{w}^{s}_{M+}$.
We train the encoder on only the pairs of sampled latent code and generated images (no real image is used) because $\bm{w}^{s}_{M+}$ corresponding to given images is unavailable.
The downsampled images $\x_{\downarrow}$ have a resolution $8\times$ larger than the feature space $\FS$.

To respect the encoded latent codes, we use a regularization that penalizes the distance between an initial latent code and a current latent code in optimization. The regularization term for $\f$ is defined as:
\begin{align}
\Lreg(\fz) = \|\f^{0} - \f\|^{2},
\end{align}
where $\f^{0}\!=\!E(\x)$. The loss prevents the latent code $\f$ from straying to far from the encoded code $\f^{0}$. 
The final objective function for GAN inversion on $\FZS$ is given by
\begin{align}
L(\fz) = \Lrecon(\fz) + \lambda_{reg} \Lreg(\fz),
\end{align}
where $\lambda_{reg}$ is a weight of the regularization loss.
We retract the latent codes $\zmpls$ to the surface of the hypersphere of radius $\sqrt{512}$ after every iteration to compute precise gradients of the latent $\zmpls$, and we update each latent code $\z_i \in \zmpls$ independently by using:
\begin{align}
  \z_i = \sqrt{512}\frac{\z_i}{|\z_i|}.
\end{align}

We optimize the latent code $\fz$ in $\FZS$ with the aid of both the feature encoder and the optimization method with regularization.
Although employing $\ZS$ space alone limits representation capacity,
extending the limited space to the $\ZPS$ space and complementing it with a highly capable $\FS$ space
lead to increase reconstruction capability without sacrificing editing quality.

\begin{figure*}
\begin{minipage}[t]{0.5\textwidth}
  \centering
    \bgroup 
    \def\arraystretch{0.2} 
    \setlength\tabcolsep{0.2pt}
    \begin{tabular}{cccccc}
\raisebox{1.6em}{target} & 
\includegraphics[width=0.17\columnwidth]{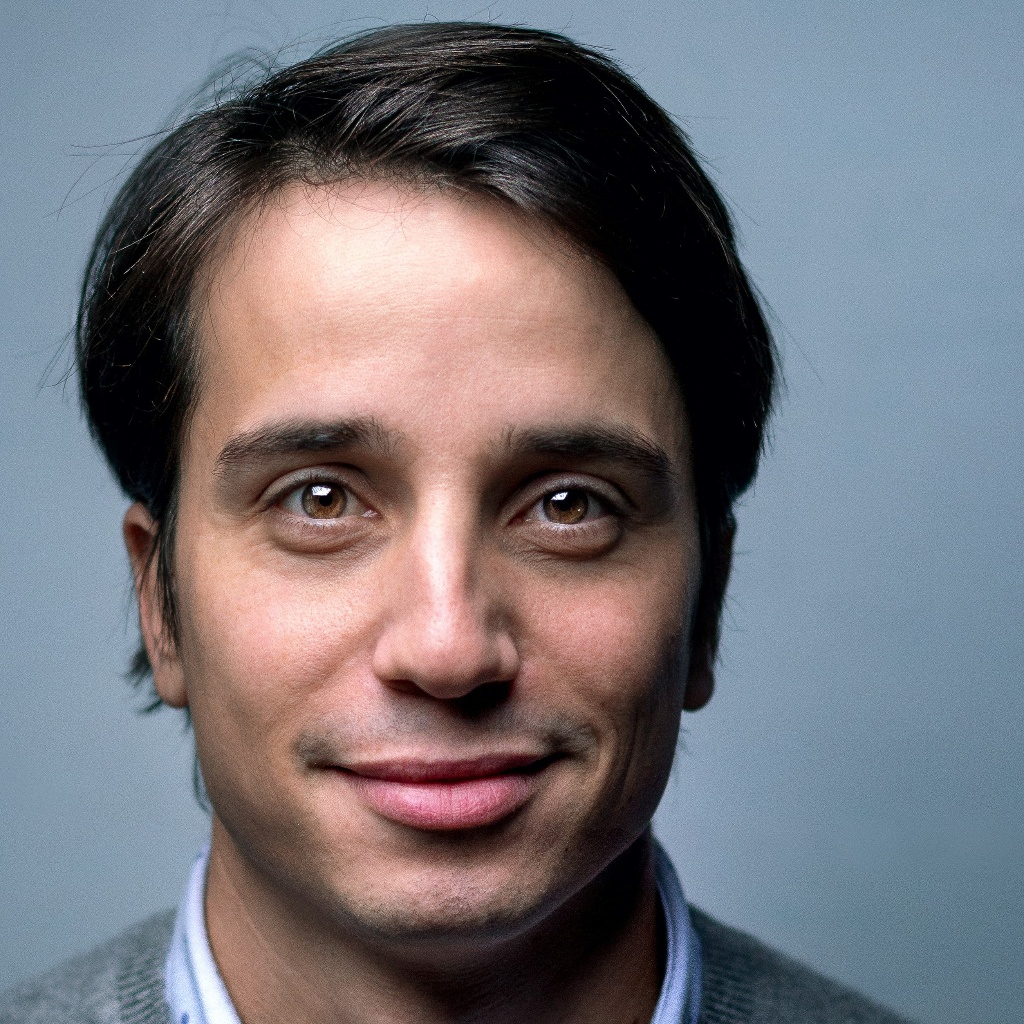} &
\includegraphics[width=0.17\columnwidth]{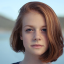} &
\includegraphics[width=0.17\columnwidth]{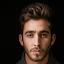} &
\includegraphics[width=0.17\columnwidth]{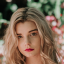} &
\includegraphics[width=0.17\columnwidth]{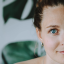} \\ \phantom{\scriptsize 0} \\\phantom{\scriptsize 0} \\\\
\raisebox{1.6em}{\begin{tabular}{c}$\mathcal{F}/\mathcal{W}^{+}$ \\ \!$(\PNS\!)$\cite{Kang_2021_ICCV}\end{tabular}} & 
\includegraphics[width=0.17\columnwidth]{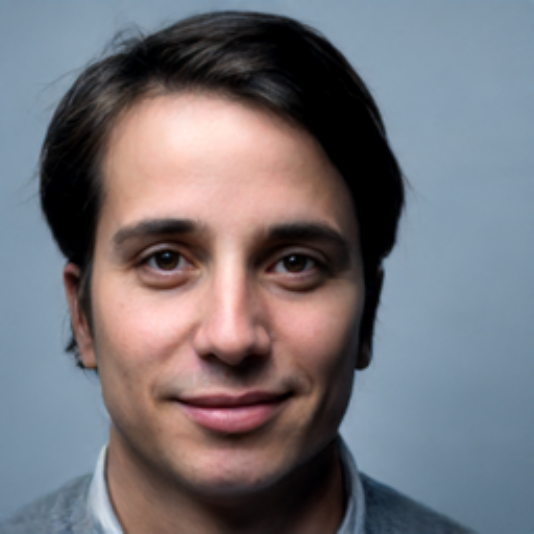} &
\includegraphics[width=0.17\columnwidth]{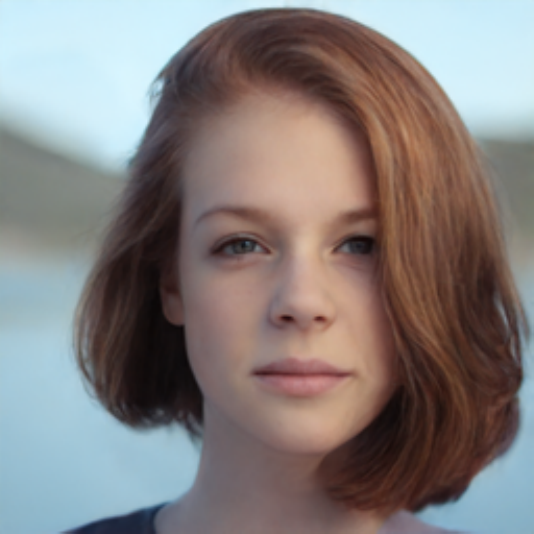} &
\includegraphics[width=0.17\columnwidth]{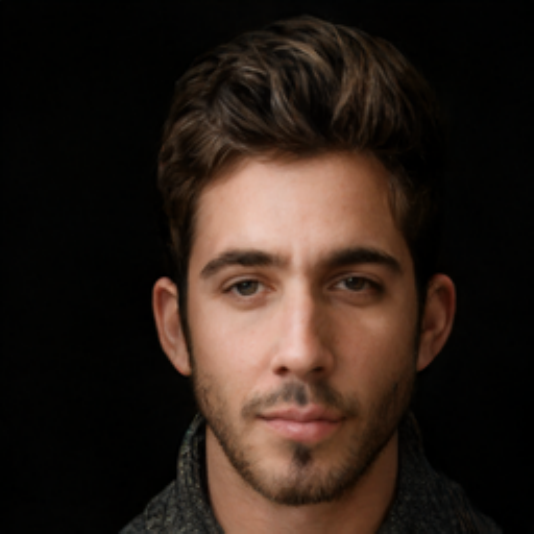} &
\includegraphics[width=0.17\columnwidth]{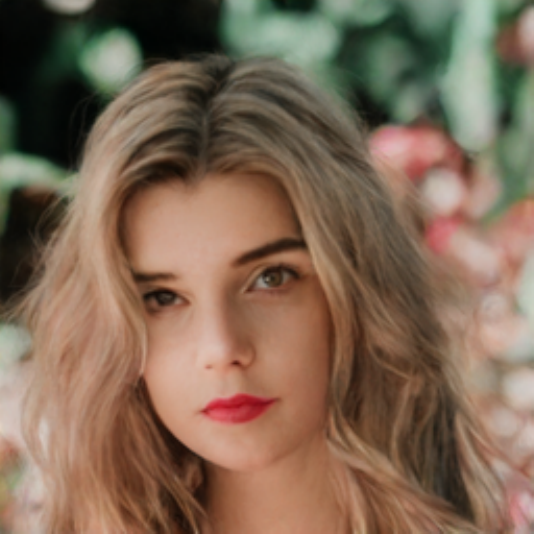} &
\includegraphics[width=0.17\columnwidth]{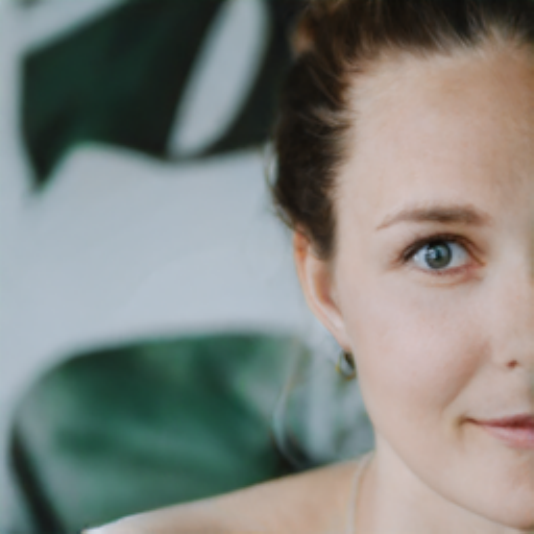} \\
\footnotesize LPIPS loss & \footnotesize 0.05944 & \footnotesize 0.06785 & \footnotesize 0.06728 & \footnotesize 0.07973 & \footnotesize 0.09912 \\
\footnotesize MSE loss & \footnotesize 0.00450 & \footnotesize 0.00680 & \footnotesize 0.01178 & \footnotesize 0.01437 & \footnotesize 0.01388 \\\\
\raisebox{1.4em}{\!$\mathcal{F}/\mathcal{S}$\!~\cite{yao2022style}\!} & 
\includegraphics[width=0.17\columnwidth]{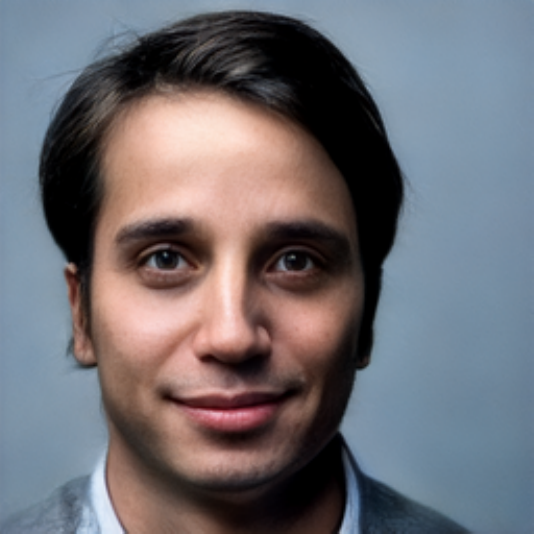} &
\includegraphics[width=0.17\columnwidth]{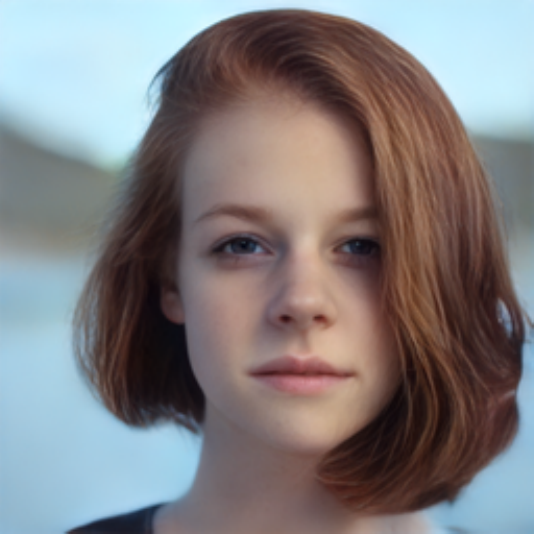} &
\includegraphics[width=0.17\columnwidth]{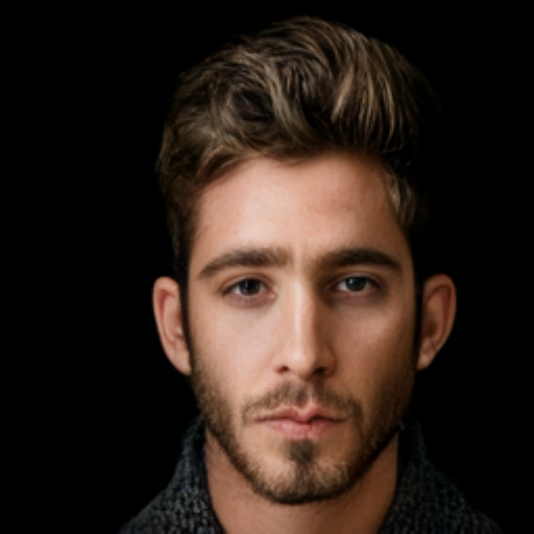} &
\includegraphics[width=0.17\columnwidth]{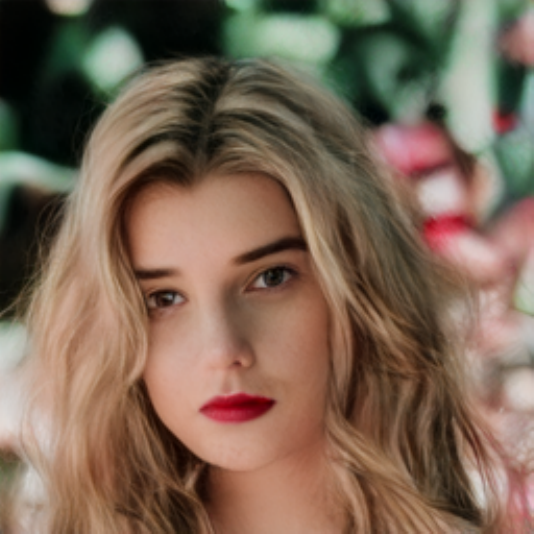} &
\includegraphics[width=0.17\columnwidth]{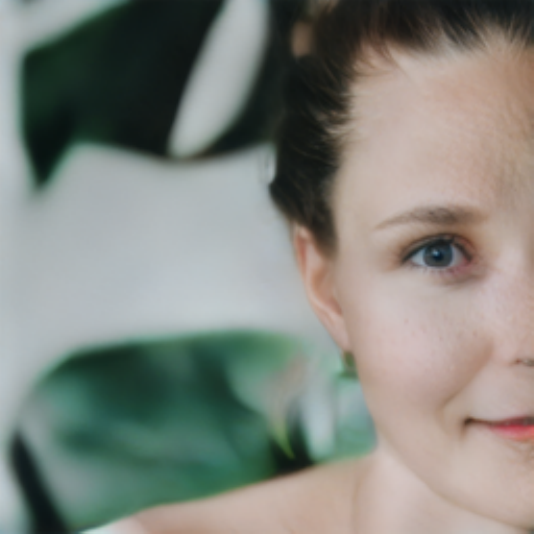} \\
\footnotesize LPIPS loss & \footnotesize 0.10868 & \footnotesize 0.07482 & \footnotesize 0.43191 & \footnotesize 0.12273 & \footnotesize $\textbf{0.07196}$ \\
\footnotesize MSE loss & \footnotesize 0.01221 & \footnotesize 0.00997 & \footnotesize $\textbf{0.01169}$ & \footnotesize 0.02023 & \footnotesize $\textbf{0.00944}$ \\\\
\raisebox{1.6em}{\shortstack{$\mathcal{F}/\mathcal{Z}$ \\ (Ours)}} & 
\includegraphics[width=0.17\columnwidth]{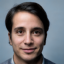} &
\includegraphics[width=0.17\columnwidth]{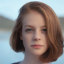} &
\includegraphics[width=0.17\columnwidth]{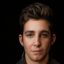} &
\includegraphics[width=0.17\columnwidth]{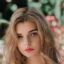} &
\includegraphics[width=0.17\columnwidth]{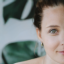} \\
\footnotesize LPIPS loss & \footnotesize 0.06573 & \footnotesize 0.07567 & \footnotesize 0.08897 & \footnotesize 0.08939 & \footnotesize 0.11237 \\
\footnotesize MSE loss & \footnotesize 0.00555 & \footnotesize 0.00994 & \footnotesize 0.01595 & \footnotesize 0.01695 & \footnotesize 0.01636 \\\\
\raisebox{1.6em}{\shortstack{$\mathcal{F}/\mathcal{Z}^{+}$ \\ (Ours)}} & 
\includegraphics[width=0.17\columnwidth]{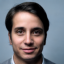} &
\includegraphics[width=0.17\columnwidth]{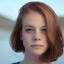} &
\includegraphics[width=0.17\columnwidth]{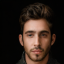} &
\includegraphics[width=0.17\columnwidth]{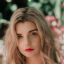} &
\includegraphics[width=0.17\columnwidth]{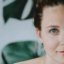} \\
\footnotesize LPIPS loss & \footnotesize $\textbf{0.05315}$ & \footnotesize $\textbf{0.06540}$ & \footnotesize $\textbf{0.06563}$ & \footnotesize $\textbf{0.07737}$ & \footnotesize 0.09158 \\
\footnotesize MSE loss & \footnotesize $\textbf{0.00479}$ & \footnotesize $\textbf{0.00673}$ & \footnotesize 0.01181 & \footnotesize $\textbf{0.01248}$ & \footnotesize 0.01248 \\\\
    \end{tabular}\egroup
\end{minipage}%
\begin{minipage}[b]{0.5\textwidth}
  \centering
    \bgroup 
    \def\arraystretch{0.2} 
    \setlength\tabcolsep{0.2pt}
    \begin{tabular}{cccccc}
    \raisebox{1.6em}{$\mathcal{W}$} & 
\includegraphics[width=0.17\columnwidth]{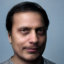} &
\includegraphics[width=0.17\columnwidth]{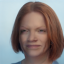} &
\includegraphics[width=0.17\columnwidth]{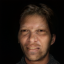} &
\includegraphics[width=0.17\columnwidth]{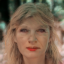} &
\includegraphics[width=0.17\columnwidth]{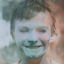} \\
\footnotesize LPIPS loss & \footnotesize 0.47981 & \footnotesize 0.26571 & \footnotesize 0.23790 & \footnotesize 0.23299 & \footnotesize 0.43533 \\
\footnotesize MSE loss & \footnotesize 0.17645 & \footnotesize 0.06406 & \footnotesize 0.06575 & \footnotesize 0.07669 & \footnotesize 0.12000 \\\\
\raisebox{1.6em}{$\mathcal{W}^{+}$} & 
\includegraphics[width=0.17\columnwidth]{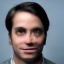} &
\includegraphics[width=0.17\columnwidth]{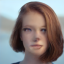} &
\includegraphics[width=0.17\columnwidth]{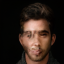} &
\includegraphics[width=0.17\columnwidth]{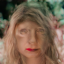} &
\includegraphics[width=0.17\columnwidth]{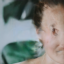} \\
\footnotesize LPIPS loss & \footnotesize 0.16998 & \footnotesize 0.11330 & \footnotesize 0.10955 & \footnotesize 0.10468 & \footnotesize 0.27122 \\
\footnotesize MSE loss & \footnotesize 0.02067 & \footnotesize 0.01873 & \footnotesize 0.02453 & \footnotesize 0.02005 & \footnotesize 0.05659 \\\\
\raisebox{1.6em}{\shortstack{\small IDInvert\\\cite{zhu2020domain}}} &
\includegraphics[width=0.17\columnwidth]{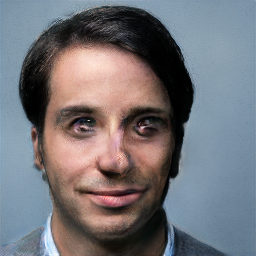} &
\includegraphics[width=0.17\columnwidth]{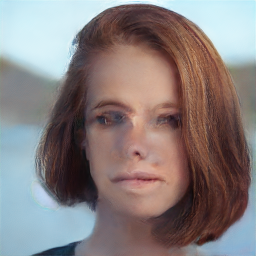} &
\includegraphics[width=0.17\columnwidth]{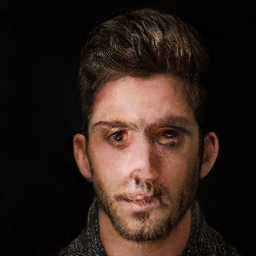} &
\includegraphics[width=0.17\columnwidth]{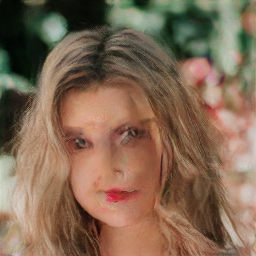} &
\includegraphics[width=0.17\columnwidth]{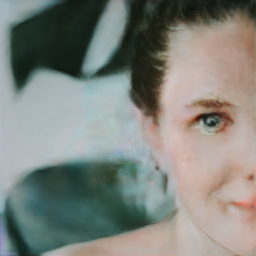} \\
\footnotesize LPIPS loss & \footnotesize 0.12845 & \footnotesize0.10068 & \footnotesize0.26865 & \footnotesize0.18732 & \footnotesize0.13559  \\
\footnotesize MSE loss & \footnotesize 0.02739 & \footnotesize0.01752 & \footnotesize0.02380 & \footnotesize0.04398 & \footnotesize0.01762  \\\\
\raisebox{1.6em}{$\mathcal{Z}$} & 
\includegraphics[width=0.17\columnwidth]{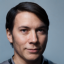} &
\includegraphics[width=0.17\columnwidth]{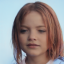} &
\includegraphics[width=0.17\columnwidth]{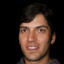} &
\includegraphics[width=0.17\columnwidth]{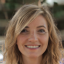} &
\includegraphics[width=0.17\columnwidth]{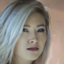} \\
\footnotesize LPIPS loss & \footnotesize 0.64010 & \footnotesize 0.39958 & \footnotesize 0.30154 & \footnotesize 0.32075 & \footnotesize 0.53057 \\
\footnotesize MSE loss & \footnotesize 0.19386 & \footnotesize 0.16928 & \footnotesize 0.08272 & \footnotesize 0.12525 & \footnotesize 0.18841 \\\\
\raisebox{1.6em}{$\mathcal{Z}^{+}$} & 
\includegraphics[width=0.17\columnwidth]{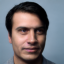} &
\includegraphics[width=0.17\columnwidth]{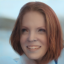} &
\includegraphics[width=0.17\columnwidth]{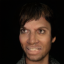} &
\includegraphics[width=0.17\columnwidth]{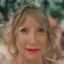} &
\includegraphics[width=0.17\columnwidth]{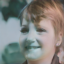} \\
\footnotesize LPIPS loss & \footnotesize 0.43211 & \footnotesize 0.25128 & \footnotesize 0.18912 & \footnotesize 0.20217 & \footnotesize 0.44184 \\
\footnotesize MSE loss & \footnotesize 0.13579 & \footnotesize 0.04729 & \footnotesize 0.05193 & \footnotesize 0.06248 & \footnotesize 0.14408 \\\\
    \end{tabular}\egroup
\end{minipage}%
    \caption{Comparison of inverted images with different latent spaces. Any latent space complemented by $\FS$ space (left column) achieves faithful reconstruction, whereas the standalone spaces (right column) fail. $\WPS$ space can achieve lower reconstruction loss with additional iterations. $\FZS$ achieves high-quality reconstructions on par $\FWS$ qualitatively, and the results of $\FZS$ are competitive with that of $\FWS$ quantitatively. Image credits are listed in Supplementary Material.}\label{fig:recon}
\end{figure*}

\begin{table*}
\centering
\caption{Quantitative comparison of latent spaces with the average MSE loss and SSIM on 50 CelebA-HQ samples. The proposed latent space $\FZS$ yields performance comparable to $\FWS(\PNS)$. }\label{tb:recon}
\begin{tabular}{ccccccccc}\hline
Space & $\mathcal{Z}$ & $\mathcal{Z}^{+}$ & $\mathcal{W}^{+} (P_N)$ & $\mathcal{W}^{+}$  & $\mathcal{F}/\mathcal{Z}$ & $\mathcal{F}/\mathcal{Z}^{+}$ (Ours)& $\mathcal{F}/\mathcal{W}^{+} (P_N)$~\cite{Kang_2021_ICCV}  & IDInvert~\cite{zhu2020domain} \\ \midrule  %
MSE & 0.18149 & 0.12117  & 0.11965 & 0.04872  & 0.02679 & $\textbf{0.01742}$ & $\textbf{0.01743}$ & 0.02155\\ %
SSIM  & 0.61155 & 0.68930  & 0.68190 & 0.76101  & 0.78965 & $\textbf{0.81477}$   & $\textbf{0.81479}$ & 0.64993 \\\hline %
\end{tabular}%
\end{table*}

\subsection{Latent editing on \texorpdfstring{$\FZS$}{F/Z+}}

For latent editing on our $\FZS$ space, we apply the discovered semantic directions to $\zmpls$.
We move the latent code $\zmpls$ to $\{\z_M + \alpha \bm{n},\z_{M+1} + \alpha \bm{n},\ldots,\z_N + \alpha \bm{n}\} $ by a latent direction $\bm{n}$ with a step size $\alpha$.
We use GANSpace~\cite{NEURIPS2020_ganspace} and InterfaceGAN~\cite{shen2020interfacegan}
for finding semantic directions.

GANSpace~\cite{NEURIPS2020_ganspace} computes semantic directions based on PCA applied in the intermediate latent space.
For semantic directions on $\ZS$, we first apply PCA to $\w = f(\z)$ because PCA cannot be applied in the prior latent space $\ZS$.
This PCA gives a basis $\bm{V}$ for $\WS$. The PCA coordinates $\bm{x}_j$ of each $\w$ are computed by $\bm{x}_j = \bm{V}^{\top}(\w_j - \bm{\mu})$
where $\bm{\mu}$ is the mean of $\w$. By employing linear regression, the latent basis vector $\bm{n}_k$ on $\ZS$ is calculated by:
\begin{align}
\bm{n}_k = \argmin \sum_{j}\|\bm{n}_{k} \bm{x}^{k}_{j} - \z_j\|^{2}.
\end{align}

InterfaceGAN~\cite{shen2020interfacegan} obtains a linear hyperplane in the latent space that can well separate the data into positive and negative groups from latent codes and corresponding attributes by using a support vector machine.
 \begin{figure*}[tbh]
  \centering
    \bgroup 
    \def\arraystretch{0.2} 
    \setlength\tabcolsep{0.2pt}
    \begin{tabular}{ccccccccc}
\includegraphics[width=0.10\linewidth]{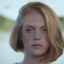} &
\includegraphics[width=0.10\linewidth]{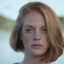} &
\includegraphics[width=0.10\linewidth]{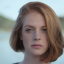} &
\includegraphics[width=0.10\linewidth]{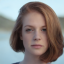} &
\includegraphics[width=0.10\linewidth]{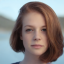} &
\includegraphics[width=0.10\linewidth]{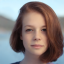} &
\includegraphics[width=0.10\linewidth]{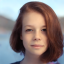} &
\includegraphics[width=0.10\linewidth]{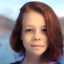} &
\includegraphics[width=0.10\linewidth]{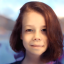} \\
\multicolumn{9}{c}{$\mathcal{F}/\mathcal{W}^{+} (P_N)$~\cite{Kang_2021_ICCV}}\\\\
\includegraphics[width=0.10\linewidth]{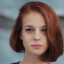} &
\includegraphics[width=0.10\linewidth]{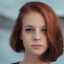} &
\includegraphics[width=0.10\linewidth]{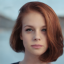} &
\includegraphics[width=0.10\linewidth]{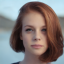} &
\includegraphics[width=0.10\linewidth]{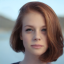} &
\includegraphics[width=0.10\linewidth]{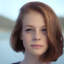} &
\includegraphics[width=0.10\linewidth]{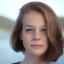} &
\includegraphics[width=0.10\linewidth]{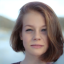} &
\includegraphics[width=0.10\linewidth]{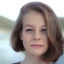} \\
\multicolumn{9}{c}{$\mathcal{F}/\mathcal{Z}^{+}$ (Ours)} 
    \end{tabular}\egroup
    \caption{Editing comparison with random directions. Editing on $\FZS$ always produces natural images while left most and right most images of $\FWS (\PNS)$ have unnatural textures.}\label{fig:random_editing}    
\end{figure*}

 \begin{figure*}[tb]
  \centering
    \bgroup 
    \def\arraystretch{0.2} 
    \setlength\tabcolsep{0.2pt}
    \begin{tabular}{cccccc|cccccc|c}
\small target & \small inversion & \multicolumn{2}{|c}{\small direction 1} & \multicolumn{2}{c|}{\small direction 2} & \small target & \small inversion & \multicolumn{2}{c}{\small direction 1} & \multicolumn{2}{c|}{\small direction 2}\\
\includegraphics[width=0.078\linewidth]{target_images/sample2} & 
\includegraphics[width=0.078\linewidth]{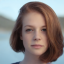} &
\includegraphics[width=0.078\linewidth]{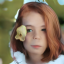} &
\includegraphics[width=0.078\linewidth]{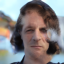} &
\includegraphics[width=0.078\linewidth]{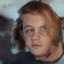} &
\includegraphics[width=0.078\linewidth]{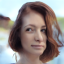} &
\includegraphics[width=0.078\linewidth]{target_images/sample4} &
\includegraphics[width=0.078\linewidth]{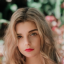} &
\includegraphics[width=0.078\linewidth]{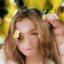} &
\includegraphics[width=0.078\linewidth]{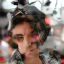} &
\includegraphics[width=0.078\linewidth]{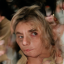} &
\includegraphics[width=0.078\linewidth]{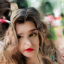} & \raisebox{1.4em}{\shortstack{\small $\mathcal{F}/\mathcal{W}^{+}$ \\$(P_N)$\cite{Kang_2021_ICCV}}} \\ &
\includegraphics[width=0.078\linewidth]{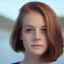} &
\includegraphics[width=0.078\linewidth]{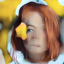} &
\includegraphics[width=0.078\linewidth]{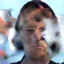} &
\includegraphics[width=0.078\linewidth]{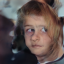} &
\includegraphics[width=0.078\linewidth]{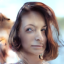} &  &
\includegraphics[width=0.078\linewidth]{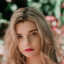} &
\includegraphics[width=0.078\linewidth]{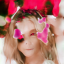} &
\includegraphics[width=0.078\linewidth]{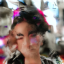} &
\includegraphics[width=0.078\linewidth]{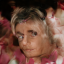} &
\includegraphics[width=0.078\linewidth]{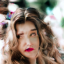} & \raisebox{1.4em}{\small $\mathcal{F}/\mathcal{W}^{+}$}\\ &
\includegraphics[width=0.078\linewidth]{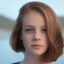} &
\includegraphics[width=0.078\linewidth]{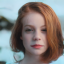} &
\includegraphics[width=0.078\linewidth]{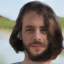} &
\includegraphics[width=0.078\linewidth]{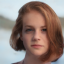} &
\includegraphics[width=0.078\linewidth]{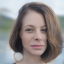} & &
\includegraphics[width=0.078\linewidth]{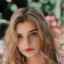} &
\includegraphics[width=0.078\linewidth]{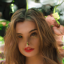} &
\includegraphics[width=0.078\linewidth]{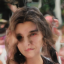} &
\includegraphics[width=0.078\linewidth]{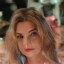} &
\includegraphics[width=0.078\linewidth]{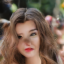} & \raisebox{1.4em}{\shortstack{\small $\mathcal{F}/\mathcal{Z}$ \\ (Ours)}} \\ &
\includegraphics[width=0.078\linewidth]{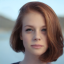} &
\includegraphics[width=0.078\linewidth]{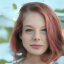} &
\includegraphics[width=0.078\linewidth]{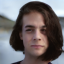} &
\includegraphics[width=0.078\linewidth]{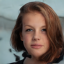} &
\includegraphics[width=0.078\linewidth]{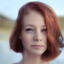} &&
\includegraphics[width=0.078\linewidth]{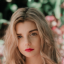} &
\includegraphics[width=0.078\linewidth]{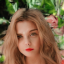} &
\includegraphics[width=0.078\linewidth]{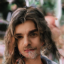} &
\includegraphics[width=0.078\linewidth]{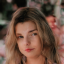} &
\includegraphics[width=0.078\linewidth]{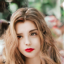} & \raisebox{1.4em}{\shortstack{\small $\mathcal{F}/\mathcal{Z}^{+}$ \\ (Ours)}}
    \end{tabular}\egroup
    \caption{Editing comparison with GANSpace directions. Although the spaces with $\WPS$ fail to preserve the structure of generated faces, the proposed spaces properly preserve them.}\label{fig:editing}
\end{figure*}

\section{Experiments}
We evaluate the proposed latent space from two aspects: reconstruction quality and editing quality.
For the reconstruction quality comparison, we verify that our latent space does not underperform the compared spaces. %
For the editing quality comparison, we demonstrate that our latent space preserves the perceptual quality of edited images better than the other spaces.

\noindent \textbf{Implementation details.}
For the inversion, we iteratively update the latent code 1200 times with the Adam optimizer.
We set learning rate of 0.01 and $\lambda_{\textrm{enc}} = \lambda_{\textrm{per}} = \lambda_{reg} = 10.0$.

\noindent \textbf{Reconstruction quality comparison.}
We first compare the reconstruction performance 
using a StyleGAN2 model pre-trained on the FFHQ dataset~\cite{Karras2019style} in both qualitative and quantitative ways.
\Cref{fig:recon} shows the reconstructed results, LPIPS loss, and MSE loss for five benchmark images on the eight compared latent spaces. We use the four latent spaces with $\FS$ (right column), four standalone spaces without $\FS$ (top four rows at left column), and IDInvert~\cite{zhu2020domain}.
Among the standalone spaces, though the $\WPS$ space performs best, it fails to reconstruct the details faithfully.
While $\ZS$ constantly produces realistic images thanks to the latent prior, it produces completely different images from the target images.
$\ZPS$ improves the reconstruction quality over $\ZS$ but is still insufficient quality.
Unlike standalone spaces, $\FS$-based latent spaces (\ie., $\FWS$, $\FSS$, $\FZS$, and $\FZ$) reconstruct well because the feature space $\FS$ magnifies the capacity of the latent space.
The LPIPS and MSE losses obtained by our $\FZS$ space are comparable to those of $\FWS$.
\Cref{fig:recon} shows that the inverted images of $\FZS$ and $\FWS$ are almost the same as the target images.

For the quantitative comparison, we prepare a set of 50 images that are randomly sampled from the CelebA-HQ dataset~\cite{karras2018progressive}. 
\Cref{tb:recon} reports the average of MSE loss and SSIM over the sampled images.
The $\FZS$ space improves the standalone spaces (\ie., $\ZS$, $\ZPS$, $\WS$, and $\WPS$) and achieves an average MSE of 0.01742 and an average SSIM of 0.81477 on the CelebA-HQ dataset. Our results are competitive with those by $\FWS(P_N)$ space, as shown by a non-inferiority test with a margin of $1\times10^{-8}$, yielding p-values of .002983 for MSE and .001609 for SSIM.
The $\FWS(P_N)$ space, however, results in less realistic edited images as seen later.

\noindent \textbf{Editing quality comparison.}
We first test whether the inverted code can move freely in the latent space using random directions to evaluate the robustness of the inverted latent code.
To this end, we first sample a 512-dimensional vector from a Gaussian distribution with a mean of 0 and a variance of 0.04. 
The intensity of editing is $[-2, 2]$.
In \cref{fig:random_editing}, latent editing on $\FWS(P_N)$ lacks details of edited images with high intensity (see right images). In contrast to $\FWS(P_N)$, latent editing on $\FZS$ preserves the reality of edited images with any intensity of editing.

\begin{figure}[tbh]
  \centering
    \bgroup 
    \def\arraystretch{0.2} 
    \setlength\tabcolsep{0.2pt}
    \begin{tabular}{cccccccc}
& target &&& target \\
\phantom{00} & \includegraphics[width=0.16\linewidth]{target_images/sample1} &\makebox[0.14\linewidth]{}&\makebox[0.16\linewidth]{}&
\includegraphics[width=0.16\linewidth]{target_images/sample3} &\makebox[0.16\linewidth]{}&\makebox[0.16\linewidth]{}\\
\end{tabular}\\
\begin{minipage}[b]{0.18\linewidth}
\begin{tabular}{cc}
&\footnotesize $\FWS$\!~\!\cite{Kang_2021_ICCV} \\
\raisebox{1.6em}{\rotatebox[origin=c]{90}{\footnotesize inversion}} &
\includegraphics[width=0.889\linewidth]{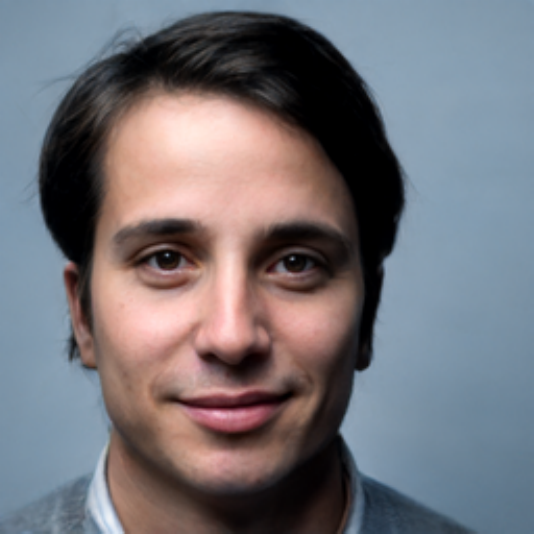} \\
\raisebox{1.7em}{\rotatebox[origin=c]{90}{\footnotesize makeup}} &
\includegraphics[width=0.889\linewidth]{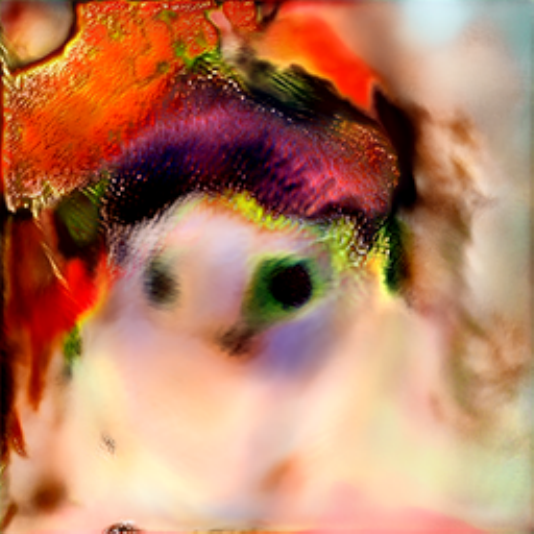} \\
\raisebox{1.8em}{\rotatebox[origin=c]{90}{\footnotesize smiling}} &
\includegraphics[width=0.889\linewidth]{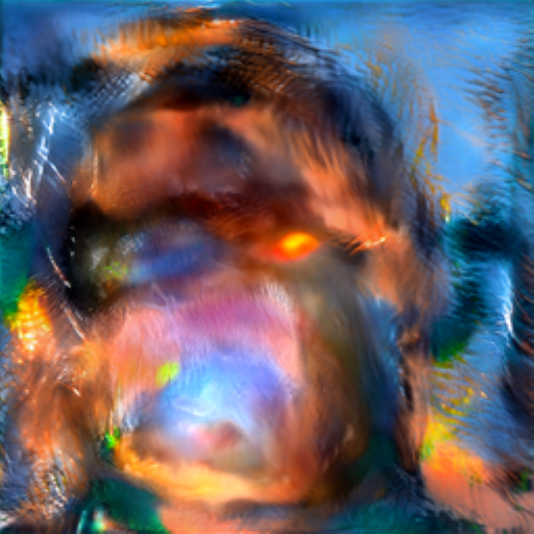} \\
\end{tabular}
\end{minipage}%
\begin{minipage}[b]{0.02\linewidth}
\end{minipage}%
\begin{minipage}[b]{0.16\linewidth}
\begin{tabular}{c}
\footnotesize $\mathcal{F}\!/\!\mathcal{S}$~\cite{yao2022style} \\
\includegraphics[width=\linewidth]{interfacegan/sp_sample1_1.pdf} \\
\includegraphics[width=\linewidth]{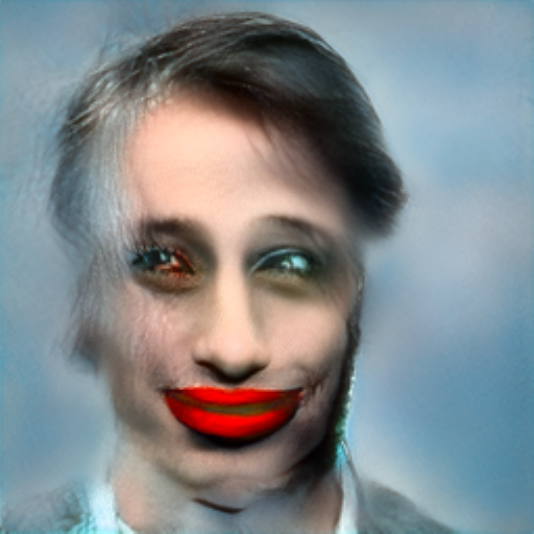} \\
\includegraphics[width=\linewidth]{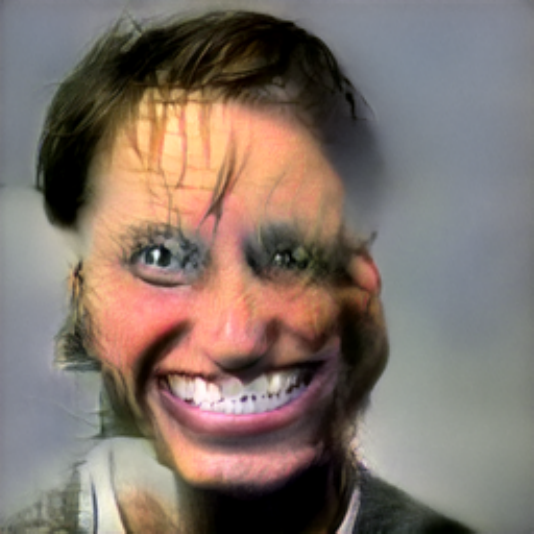} \\
\end{tabular}
\end{minipage}%
\begin{minipage}[b]{0.02\linewidth}
\end{minipage}%
\begin{minipage}[b]{0.16\linewidth}
\begin{tabular}{c}
\footnotesize $\FZS$ \\
\includegraphics[width=\linewidth]{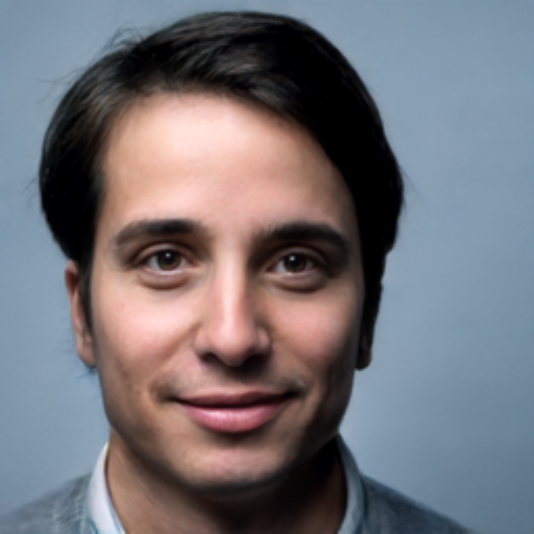} \\
\includegraphics[width=\linewidth]{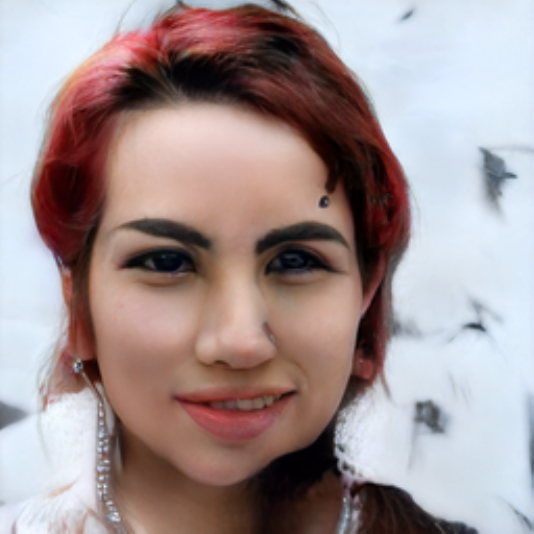} \\
\includegraphics[width=\linewidth]{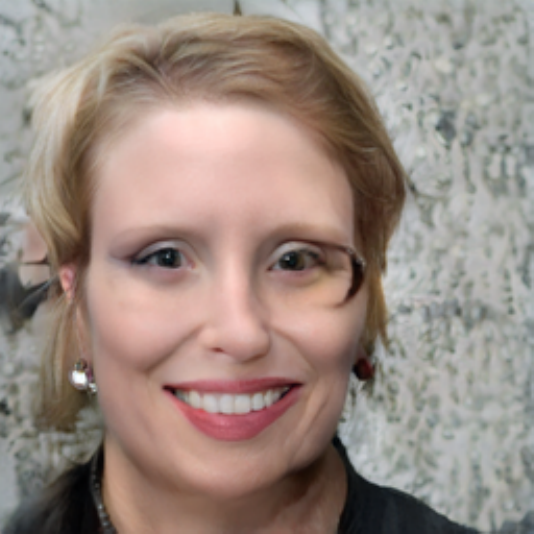} \\
\end{tabular}
\end{minipage}%
\begin{minipage}[b]{0.02\linewidth}
\end{minipage}%
\begin{minipage}[b]{0.16\linewidth}
\begin{tabular}{c}
\footnotesize $\FWS$\!~\!\cite{Kang_2021_ICCV} \\
\includegraphics[width=\linewidth]{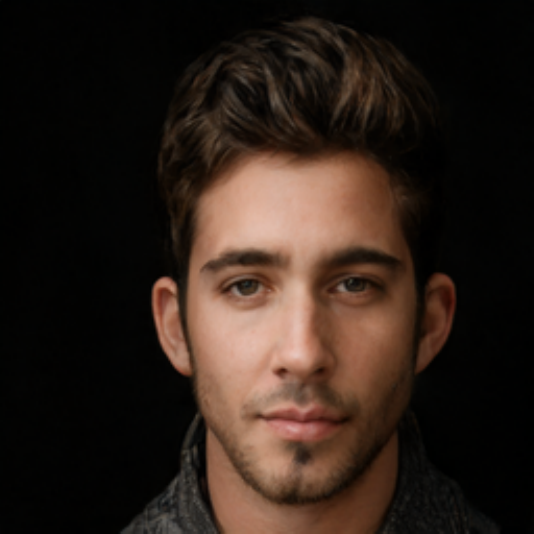} \\
\includegraphics[width=\linewidth]{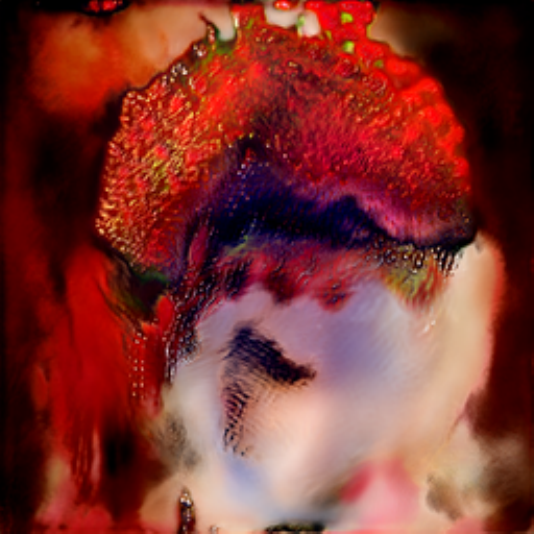} \\
\includegraphics[width=\linewidth]{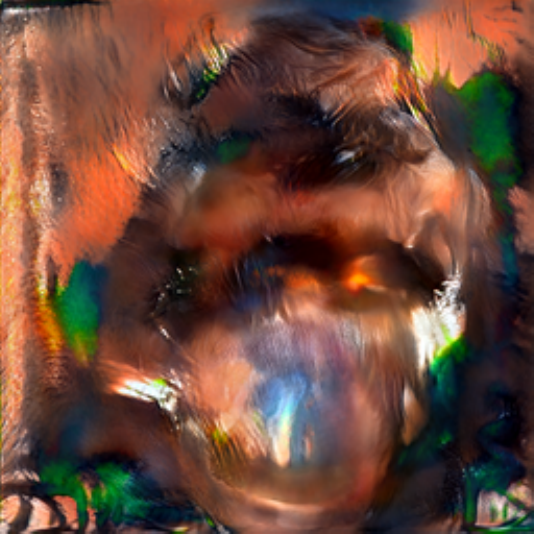} \\
\end{tabular}
\end{minipage}%
\begin{minipage}[b]{0.02\linewidth}
\end{minipage}%
\begin{minipage}[b]{0.16\linewidth}
\begin{tabular}{c}
\footnotesize $\mathcal{F}\!/\!\mathcal{S}$~\cite{yao2022style} \\
\includegraphics[width=\linewidth]{interfacegan/sp_sample3_1.pdf} \\
\includegraphics[width=\linewidth]{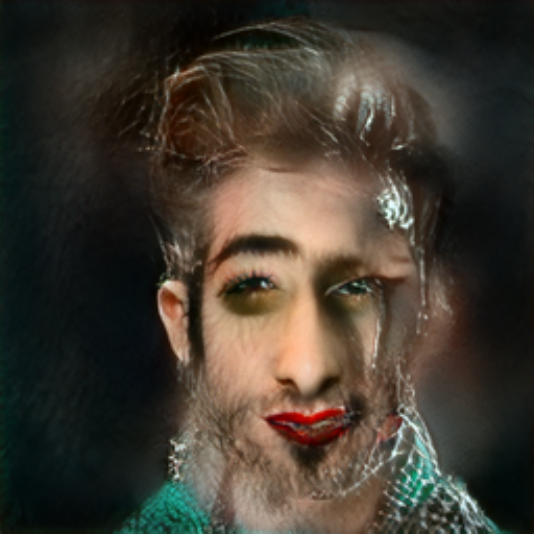} \\
\includegraphics[width=\linewidth]{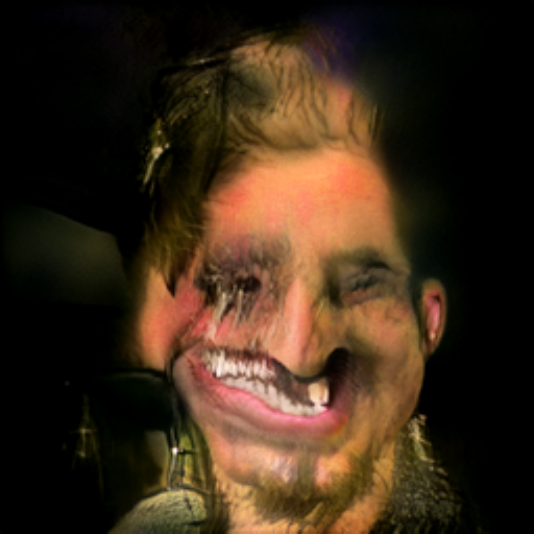} \\
\end{tabular}
\end{minipage}%
\begin{minipage}[b]{0.02\linewidth}
\end{minipage}%
\begin{minipage}[b]{0.16\linewidth}
\begin{tabular}{c}
\footnotesize $\FZS$ \\
\includegraphics[width=\linewidth]{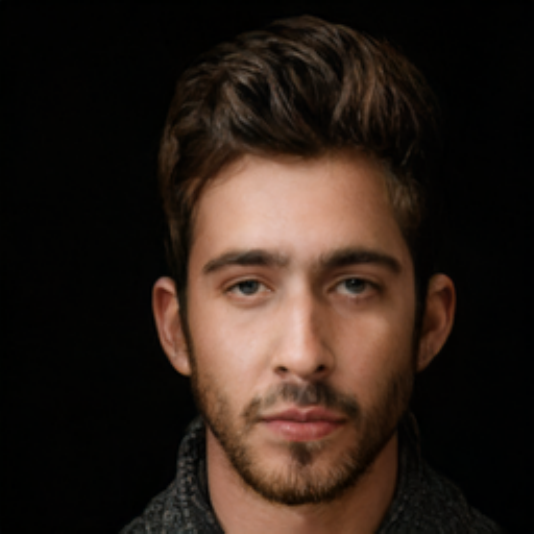} \\
\includegraphics[width=\linewidth]{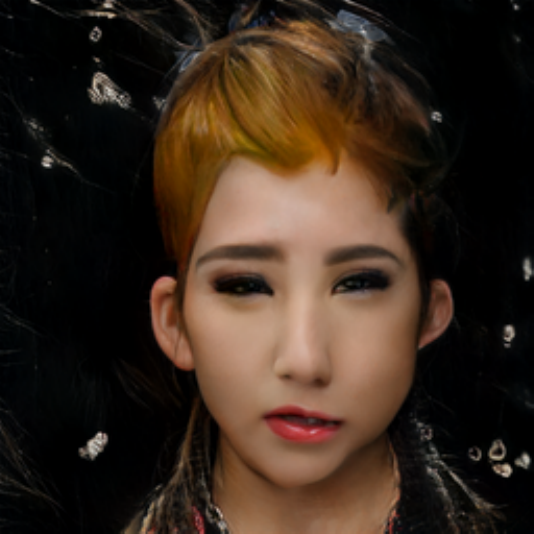} \\
\includegraphics[width=\linewidth]{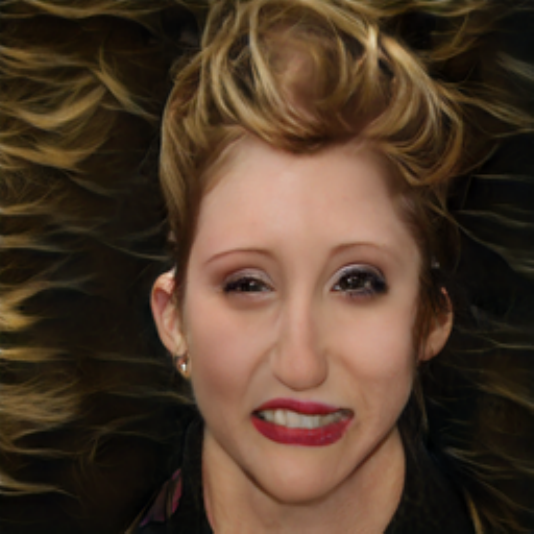} \\
\end{tabular}
\end{minipage}%
\egroup
\caption{Editing comparison with directions obtained by InterfaceGAN~\cite{shen2020interfacegan}.}\label{fig:interfacegan}
\end{figure}

\begin{figure}[tb]
\centering
\includegraphics[width=0.95\columnwidth]{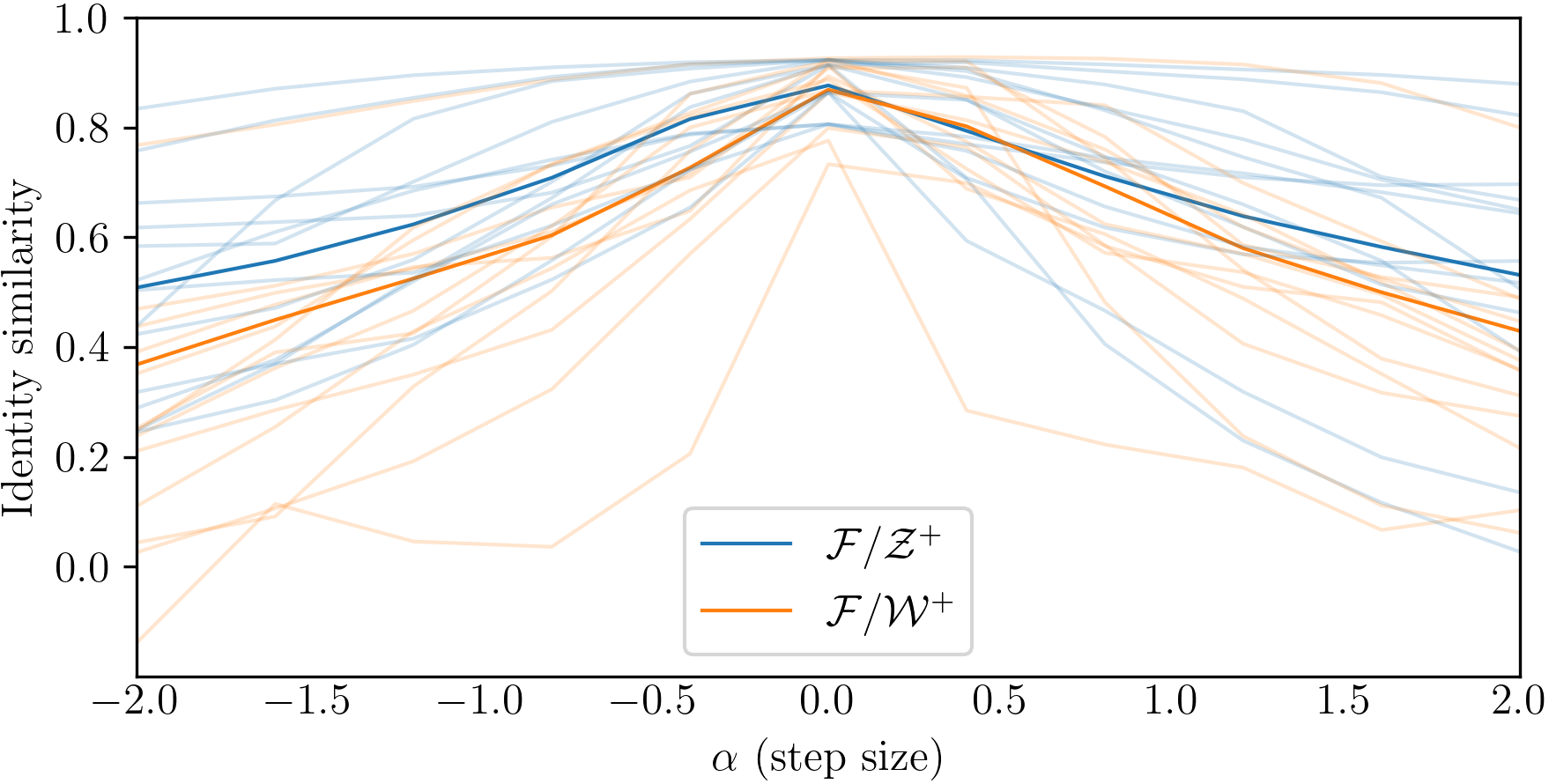}
\caption{Quantitative metric for evaluating editing quality. We compute the identity similarity between the target images and the edited images. Each light-colored line indicates the results of a semantic direction. Each deep-colored line indicates the mean of the results of each method. Achieving high identity similarity in cases with strong intensity indicates that the proposed method has higher editing quality than the compared method.}\label{fig:identity_similarity}
\end{figure}

\begin{figure}[tbh]
  \centering
    \bgroup 
    \def\arraystretch{0.2} 
    \setlength\tabcolsep{0.2pt}
    \begin{tabular}{ccccccc}
    & \multicolumn{6}{c}{2nd direction} \\
\raisebox{1.6em}{\footnotesize \begin{tabular}{c}$\FWS$ \\ $(\PNS)$\end{tabular}} & 
\includegraphics[width=0.145\columnwidth]{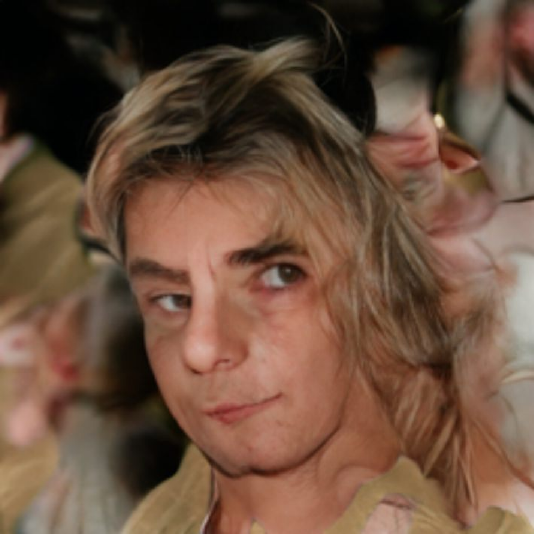} &
\includegraphics[width=0.145\columnwidth]{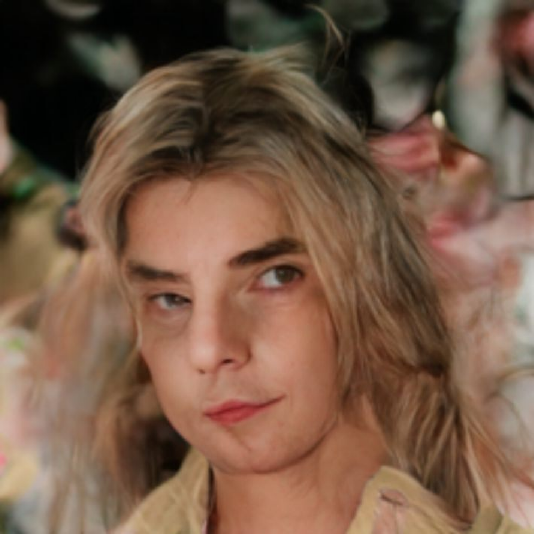} &
\includegraphics[width=0.145\columnwidth]{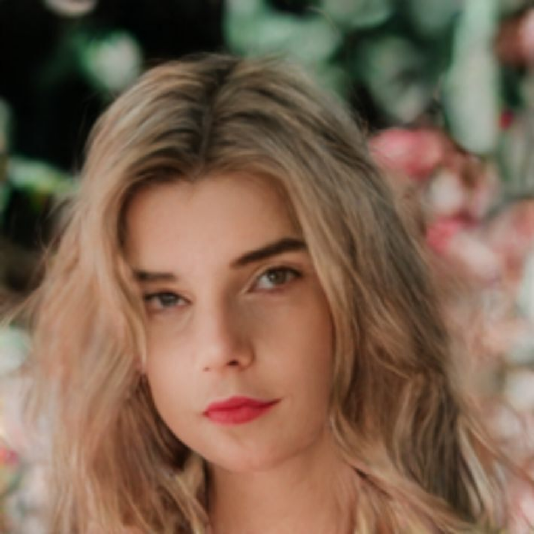} &
\includegraphics[width=0.145\columnwidth]{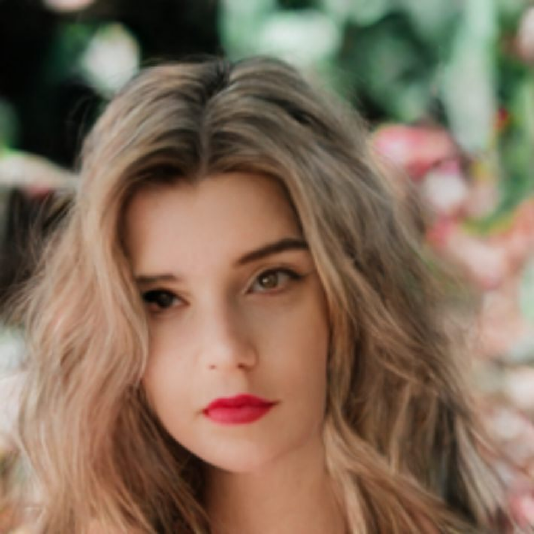} &
\includegraphics[width=0.145\columnwidth]{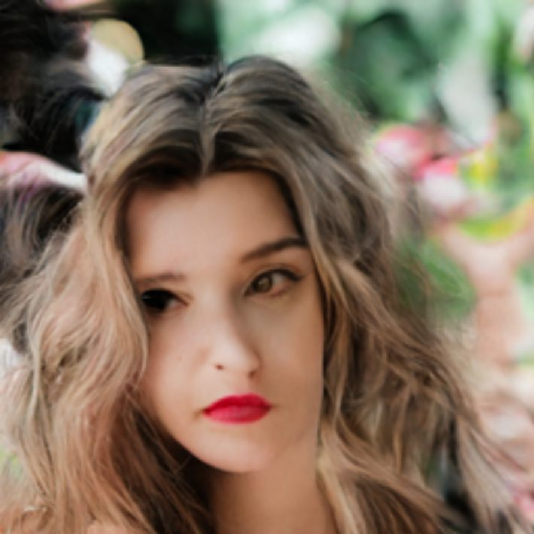} &
\includegraphics[width=0.145\columnwidth]{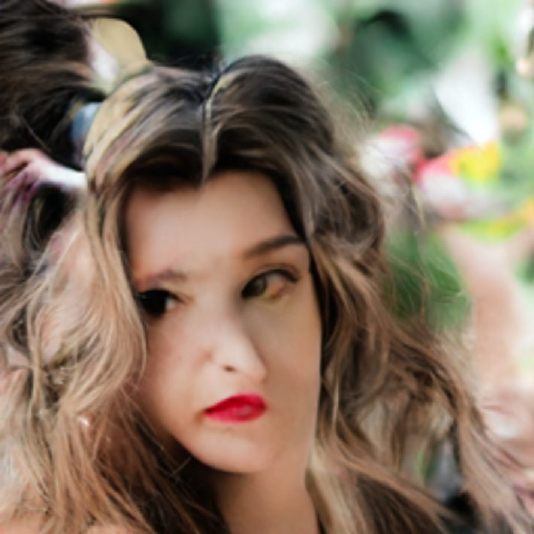} \\
\raisebox{1.4em}{\footnotesize $\FZS$} & 
\includegraphics[width=0.145\columnwidth]{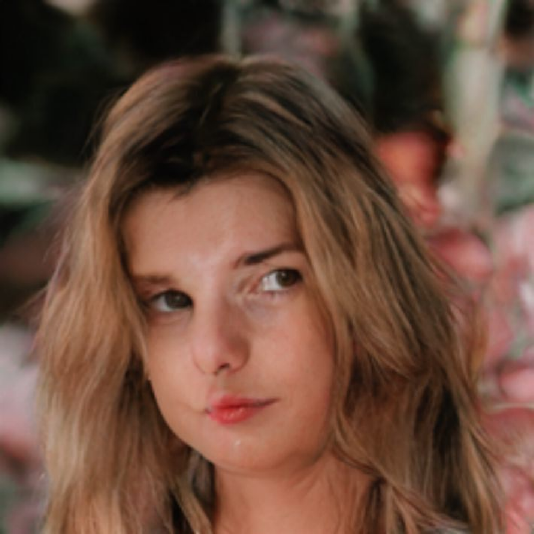} &
\includegraphics[width=0.145\columnwidth]{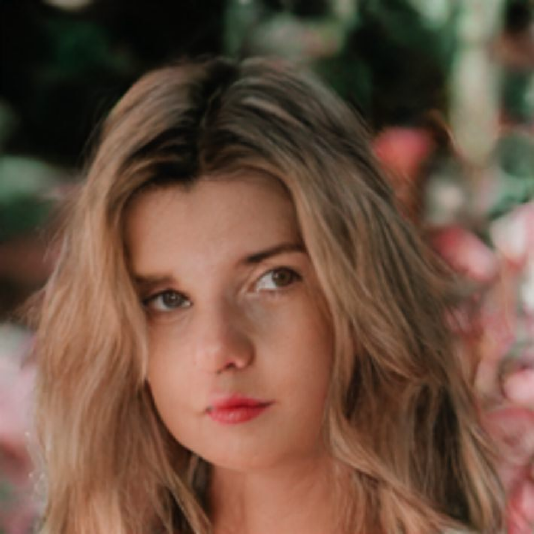} &
\includegraphics[width=0.145\columnwidth]{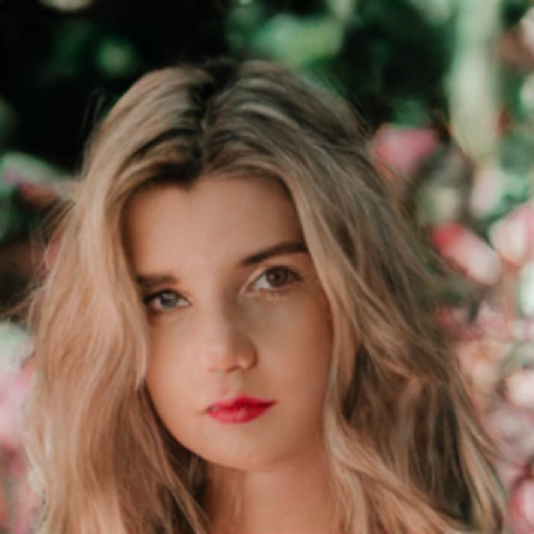} &
\includegraphics[width=0.145\columnwidth]{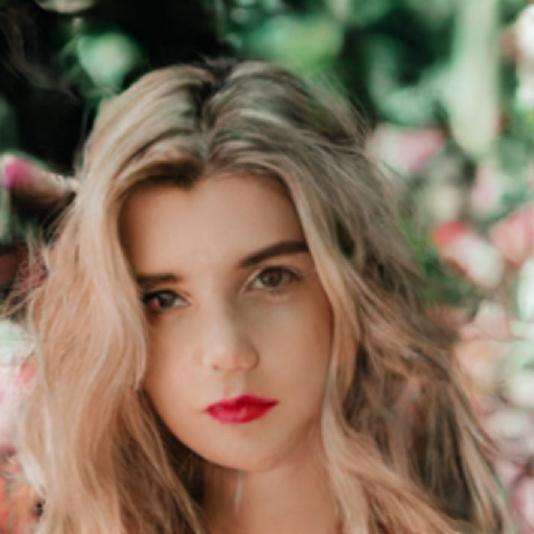} &
\includegraphics[width=0.145\columnwidth]{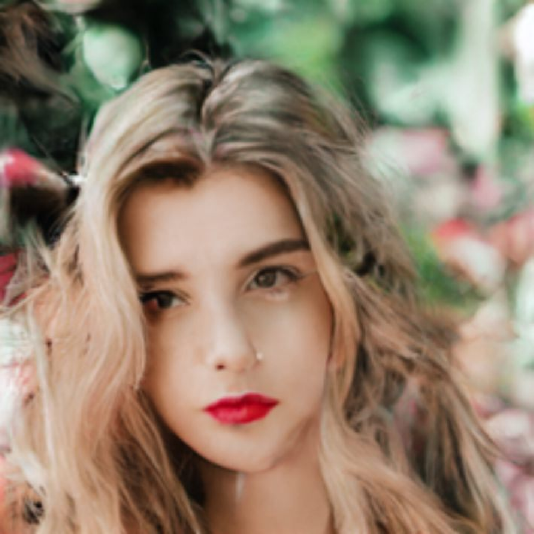} &
\includegraphics[width=0.145\columnwidth]{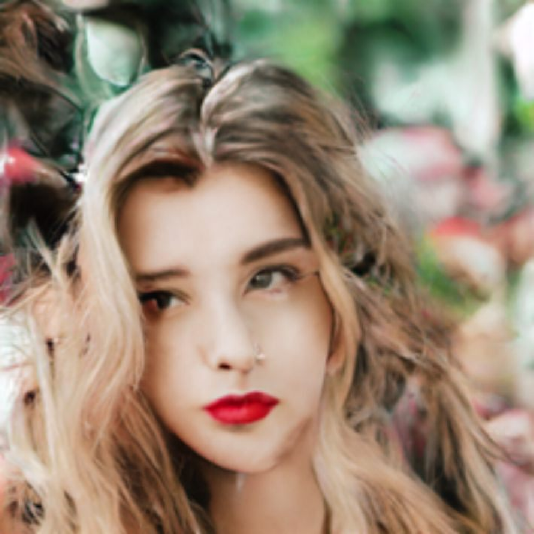} \\
& \multicolumn{6}{c}{\footnotesize -2.0 $\xlongleftarrow[]{\hspace{7em}}$ step size $\xlongrightarrow[]{\hspace{7em}}$ 2.0}\\\\
    & \multicolumn{6}{c}{3rd direction} \\
\raisebox{1.6em}{\footnotesize \begin{tabular}{c}$\FWS$ \\ $(\PNS)$\end{tabular}} & 
\includegraphics[width=0.145\columnwidth]{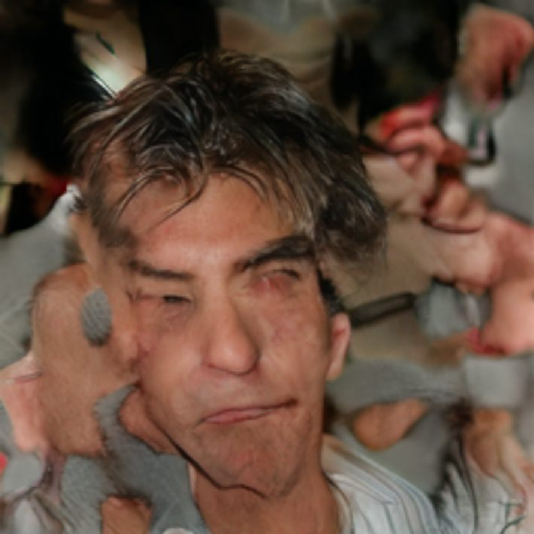} &
\includegraphics[width=0.145\columnwidth]{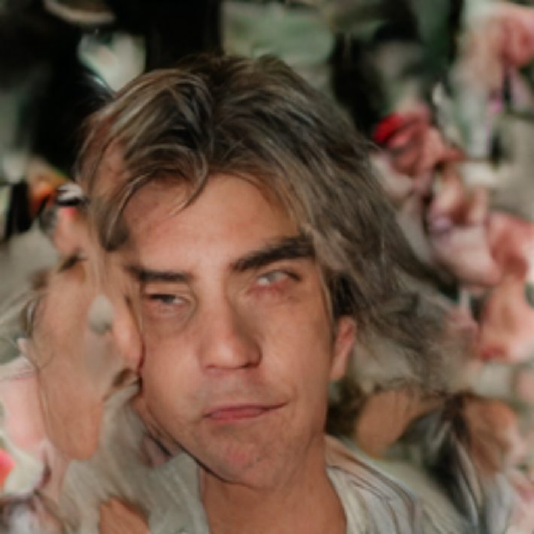} &
\includegraphics[width=0.145\columnwidth]{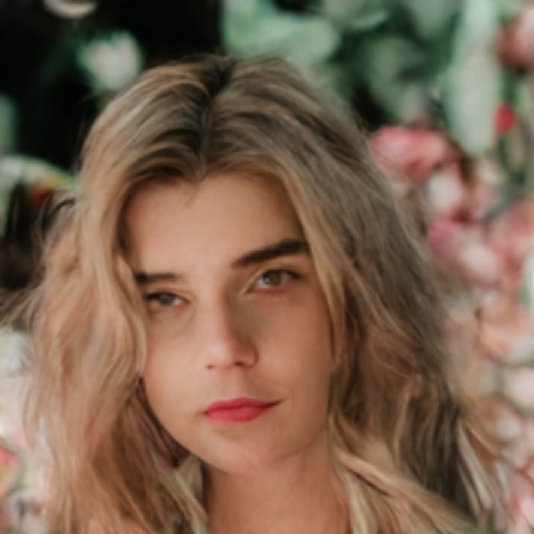} &
\includegraphics[width=0.145\columnwidth]{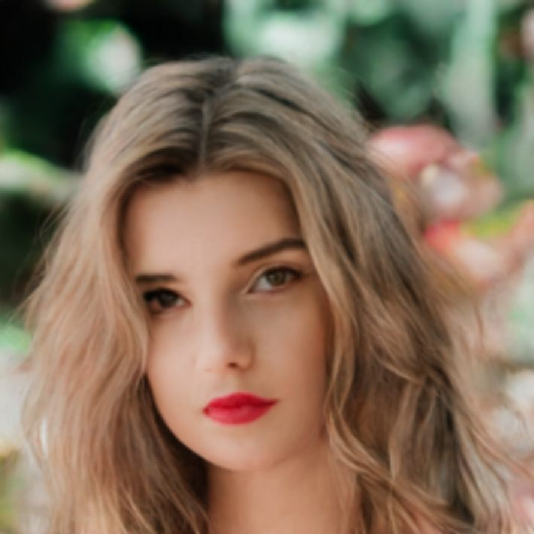} &
\includegraphics[width=0.145\columnwidth]{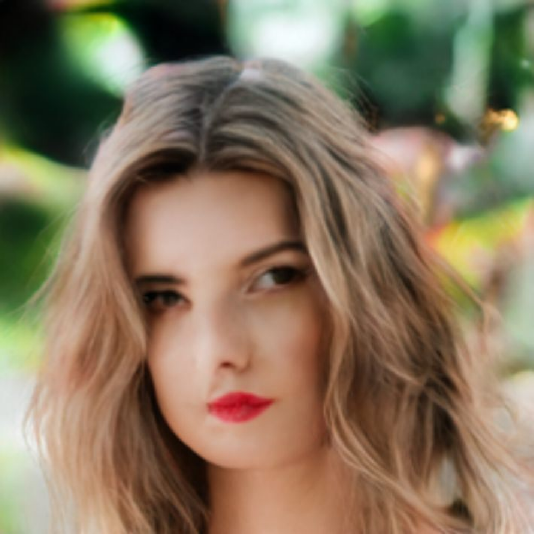} &
\includegraphics[width=0.145\columnwidth]{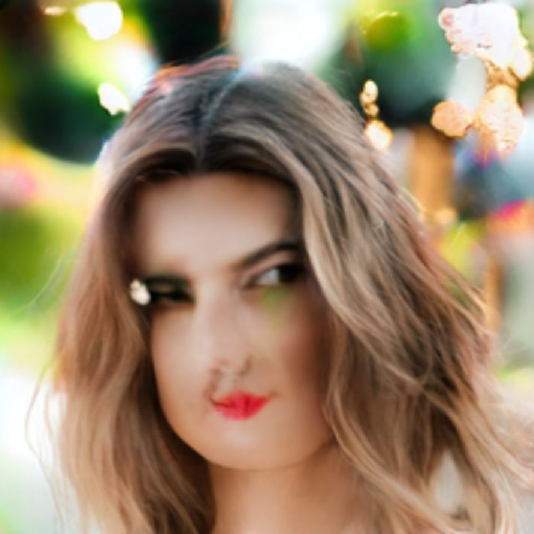} \\
\raisebox{1.4em}{\footnotesize $\FZS$} & 
\includegraphics[width=0.145\columnwidth]{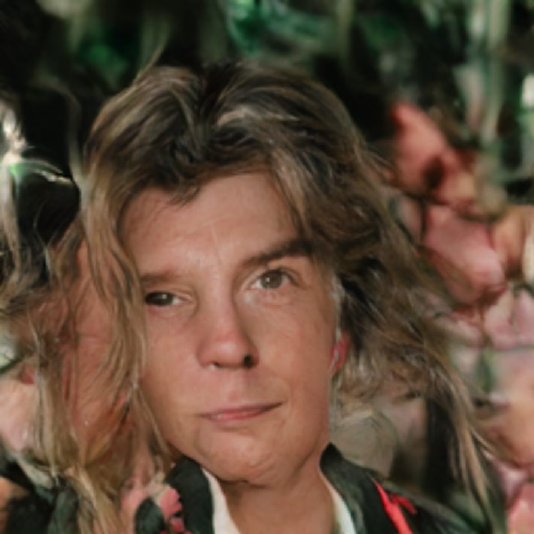} &
\includegraphics[width=0.145\columnwidth]{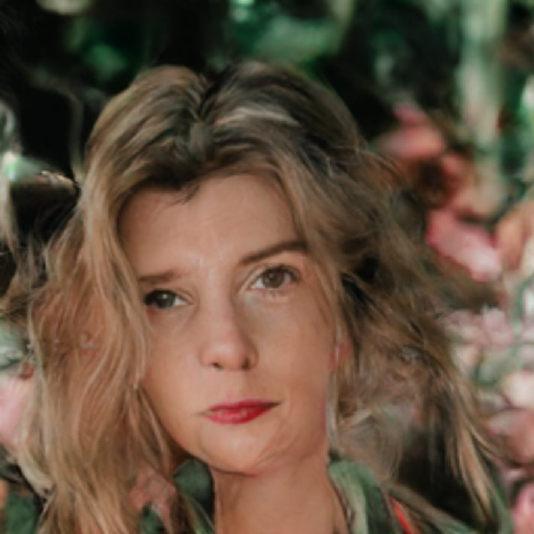} &
\includegraphics[width=0.145\columnwidth]{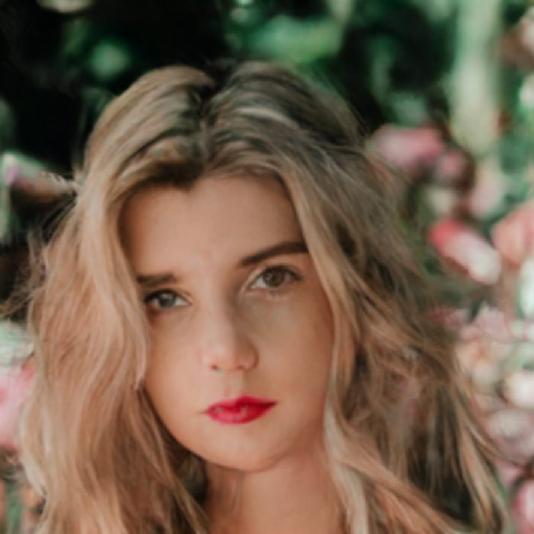} &
\includegraphics[width=0.145\columnwidth]{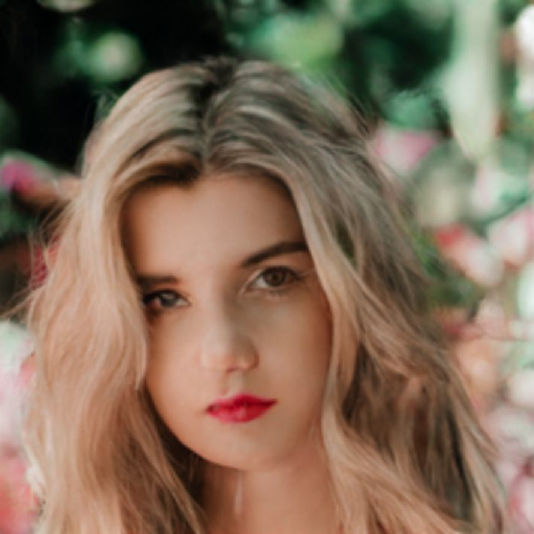} &
\includegraphics[width=0.145\columnwidth]{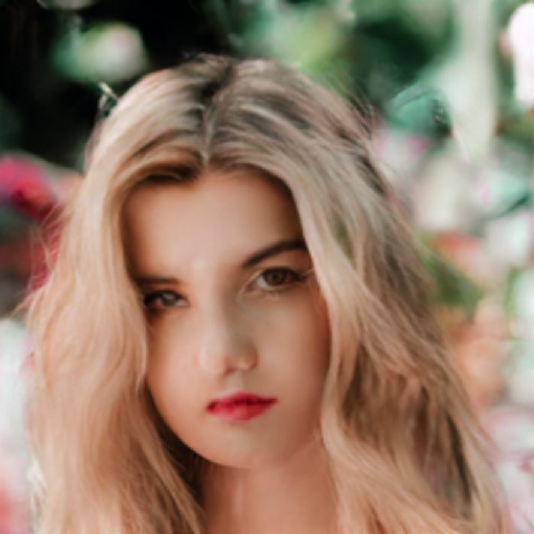} &
\includegraphics[width=0.145\columnwidth]{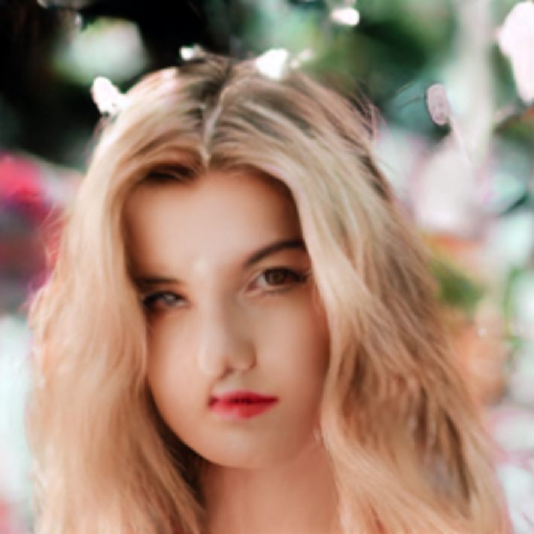} \\
& \multicolumn{6}{c}{\footnotesize -2.0 $\xlongleftarrow[]{\hspace{7em}}$ step size $\xlongrightarrow[]{\hspace{7em}}$ 2.0}\\\\
    & \multicolumn{6}{c}{ 4th direction} \\
\raisebox{1.6em}{\footnotesize \begin{tabular}{c}$\FWS$ \\ $(\PNS)$\end{tabular}} & 
\includegraphics[width=0.145\columnwidth]{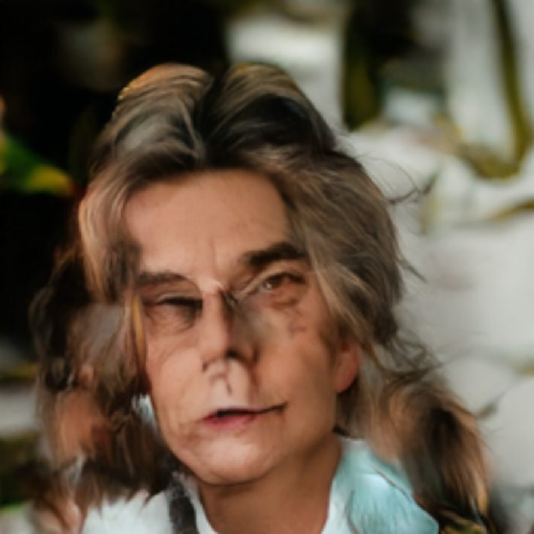} &
\includegraphics[width=0.145\columnwidth]{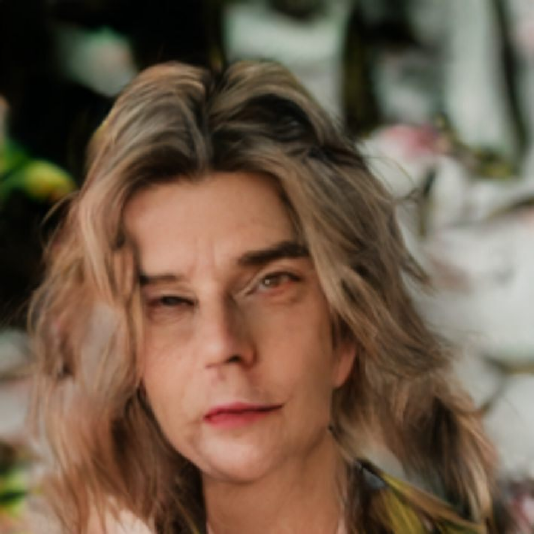} &
\includegraphics[width=0.145\columnwidth]{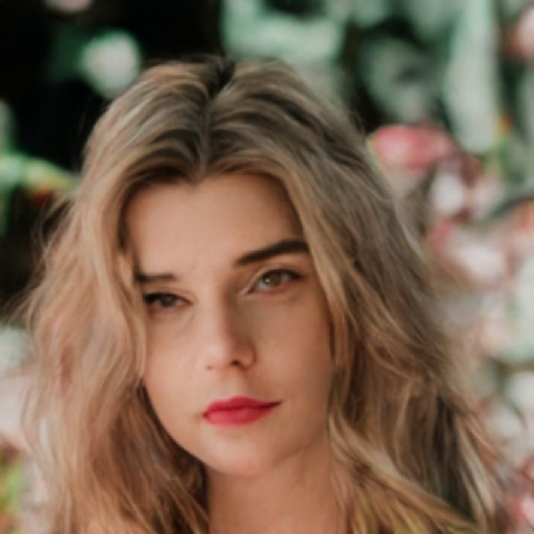} &
\includegraphics[width=0.145\columnwidth]{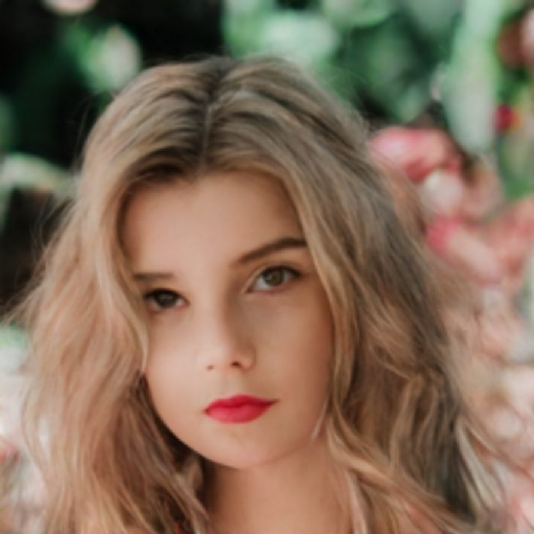} &
\includegraphics[width=0.145\columnwidth]{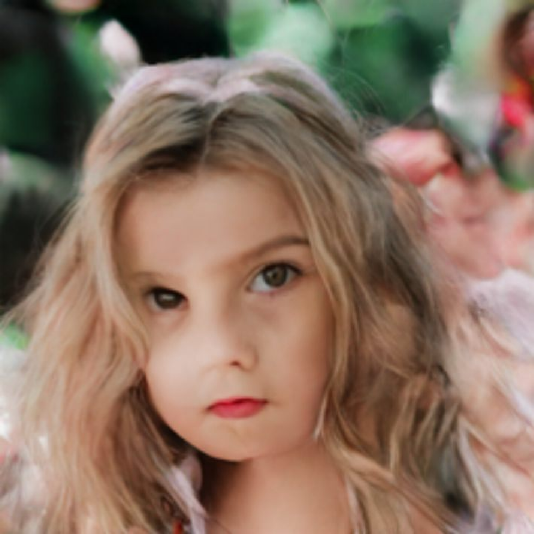} &
\includegraphics[width=0.145\columnwidth]{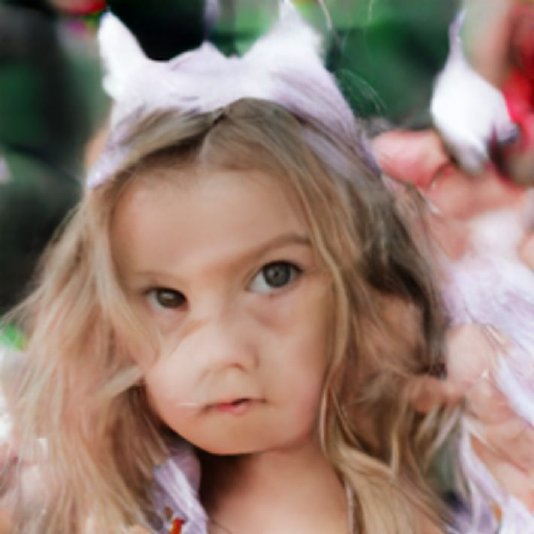} \\
\raisebox{1.4em}{\footnotesize $\FZS$} & 
\includegraphics[width=0.145\columnwidth]{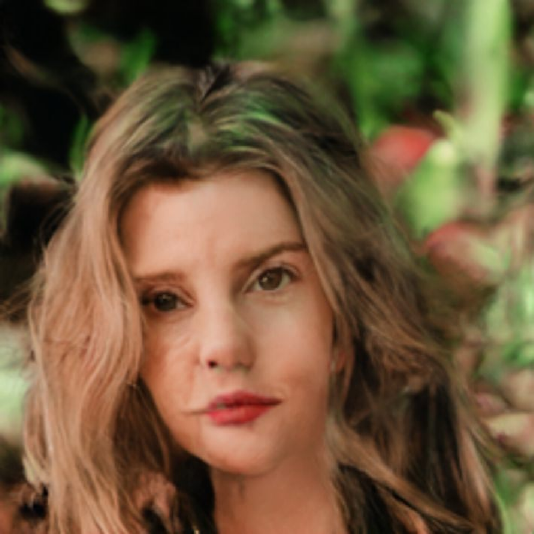} &
\includegraphics[width=0.145\columnwidth]{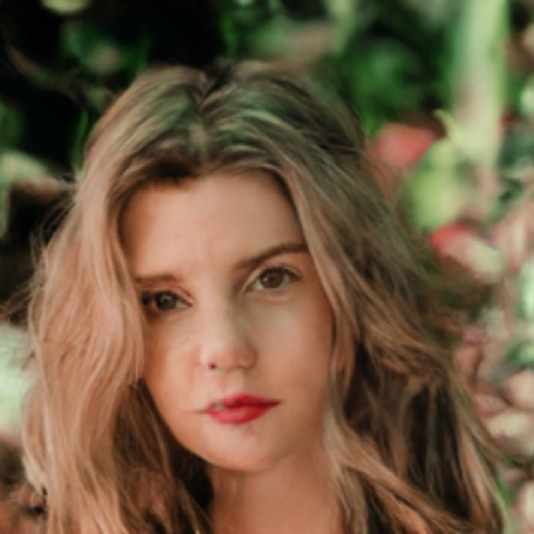} &
\includegraphics[width=0.145\columnwidth]{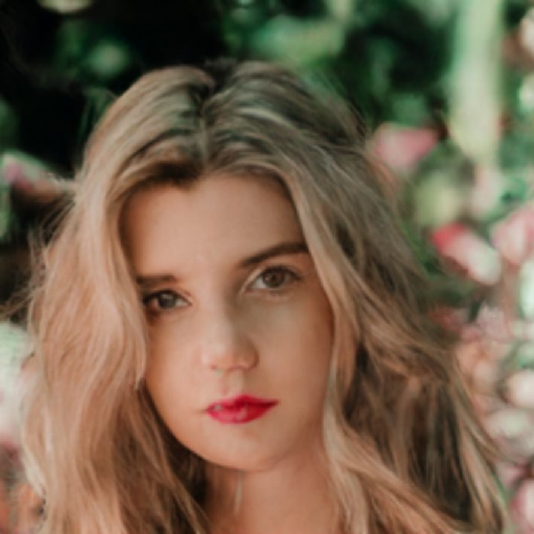} &
\includegraphics[width=0.145\columnwidth]{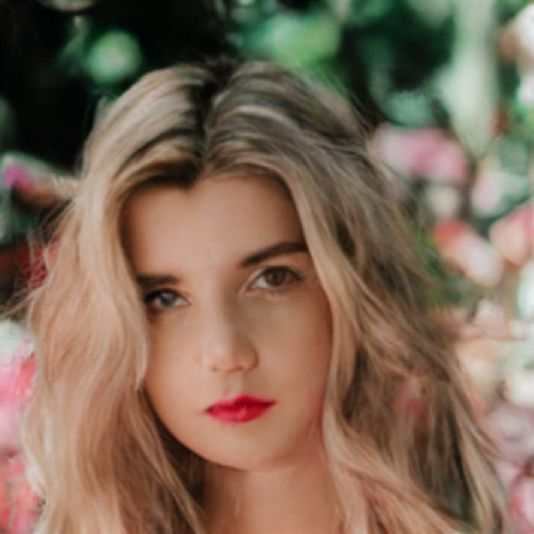} &
\includegraphics[width=0.145\columnwidth]{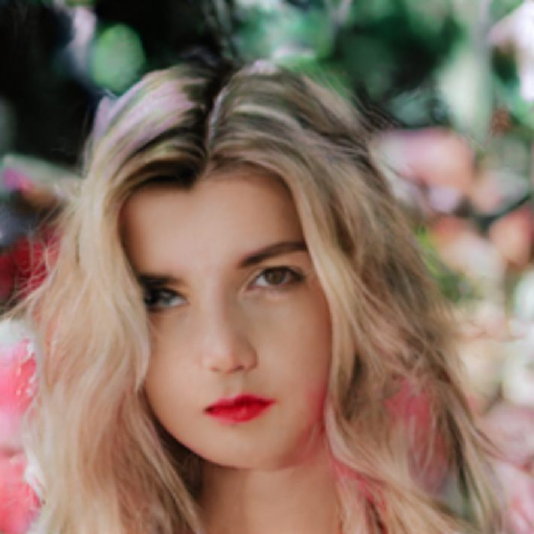} &
\includegraphics[width=0.145\columnwidth]{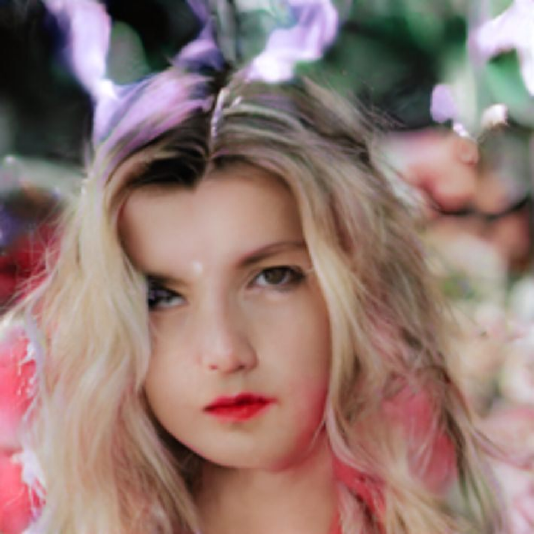} \\
& \multicolumn{6}{c}{\footnotesize -2.0 $\xlongleftarrow[]{\hspace{7em}}$ step size $\xlongrightarrow[]{\hspace{7em}}$ 2.0}
\end{tabular}\egroup
\caption{Results of the stepwise editing.
Whereas $\FWS(\PNS)$ space losses image fidelity with strong editing, our space does not lose image fidelity.
}\label{fig:stepwise_editing}
\end{figure}

\begin{figure}[tbh]
  \centering
    \bgroup 
    \def\arraystretch{0.2} 
    \setlength\tabcolsep{0.2pt}
    \begin{tabular}{cccccc}
      inversion & edited & &&inversion& edited \\
\includegraphics[width=0.18\columnwidth]{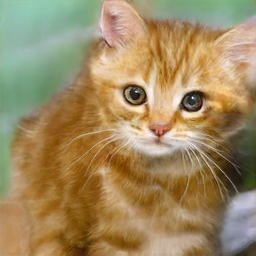} &
\includegraphics[width=0.18\columnwidth]{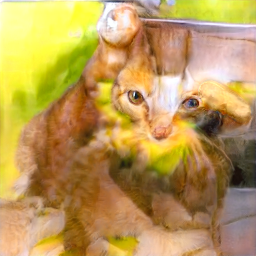} &\phantom{mmmp}&&
\includegraphics[width=0.18\columnwidth]{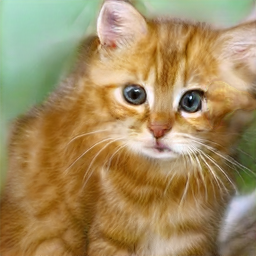} &
\includegraphics[width=0.18\columnwidth]{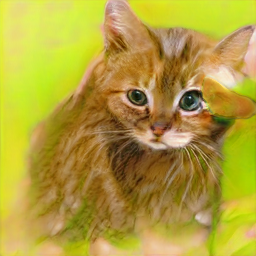} \\
\multicolumn{2}{c}{BDInvert~\cite{Kang_2021_ICCV}} &&&
\multicolumn{2}{c}{$\FZS$ (Ours)} 
\end{tabular}\egroup
    \caption{Edited results on the LSUN Cat dataset with StyleGAN1. Editing on our $\FZS$ space
      preserves the image content (cat).
    }\label{fig:cat_stylegan1}
\end{figure}

We next edit the images using actual semantic directions.
\Cref{fig:editing} shows the target images, inverted images, and edited images on the four spaces.
We use GANSpace~\cite{NEURIPS2020_ganspace} as a method of discovering semantic directions and use two directions for editing.
For each direction, two images with intensities of -2 and 2 are plotted, leading to eight edited images per method.
Although $\FWS(\PPNS)$ is more robust than $\FWS$ due to the regularization on $\PNS$ as discussed in Kang \etal\cite{Kang_2021_ICCV},
both $\FWS(P_N)$ and $\FWS$ lack the image quality after performing editing operation with GANSpace directions. 
Some images lack face parts or are added waterdrops.
Meanwhile, the proposed $\FZS$ and $\FZ$ spaces 
consistently preserve image quality after semantic editing.
We also compare editing results with interfaceGAN directions. As shown in \cref{fig:interfacegan}, we can see that our method relaxes
the distortions of edited images more than the competing methods.

In addition to \cref{fig:editing} which shows the editing results with only two step sizes, we also show the results with more step sizes and additional semantic directions in \cref{fig:stepwise_editing}.
\Cref{fig:stepwise_editing} shows the results with six step sizes in [-2.0, 2.0] and 2nd, 3rd, and 4th GANSpace directions. The $\FZS$ space maintains the image fidelity after editing.

Finally, we compare the editing quality of different spaces quantitatively.
We use MTCNN~\cite{zhang2016joint} as the face detector and InceptionResNet V1~\cite{szegedy2017inception} trained on VGGFace2~\cite{cao2018vggface2} as the feature extractor.
To compute identity similarity, we use cosine similarity. For each method, we plot the identity similarities between the original inputs and edited images with each step size of editing in \cref{fig:identity_similarity}. 
We plot 12 lines for each method (four targets $\times$ three semantic directions).
The figure shows that the proposed space preserves the identity of target images after editing with even a strong intensity unlike $\FWS(\PNS)$.
We also conduct the editing quality evaluation on 50 CelebA-HQ samples with five directions and eleven step sizes. The average identity similarity of our proposed space is 0.373, whereas that of the $\FWS$ space is 0.327.
The results demonstrate that $\FZS$ maintains the perceptual quality well.

\noindent \textbf{Editing comparison on another dataset.}
We evaluate the effectiveness of the proposed space on another GAN model.
\Cref{fig:cat_stylegan1} shows the edited results with StyleGAN 1 pretrained on the LSUN Cat dataset.
Although BDInvert completely corrupts the cat's face in the edited image, our method maintains it. We provide further inversion and editing results on additional datasets in the supplement.

Through the above comparisons, the proposed space $\FZS$ has the advantage over the existing spaces. 
Due to the utilization of the hypersphere space, $\FZS$ has higher editing quality than $\FWS$ and $\FWS (\PNS)$ without
sacrificing reconstruction performance. In fact, $\FZS$ achieves the quantitative performance comparable to $\FWS (\PNS)$, 
and the inverted results of $\FZS$ and $\FWS$ spaces are almost indistinguishable. 

\begin{figure}[t]
  \centering
    \bgroup 
    \def\arraystretch{0.2} 
    \setlength\tabcolsep{0.2pt}
    \begin{tabular}{cccccc}
\raisebox{1.5em}{\footnotesize PTI~\cite{roich2021pivotal}} &
\includegraphics[width=0.17\columnwidth]{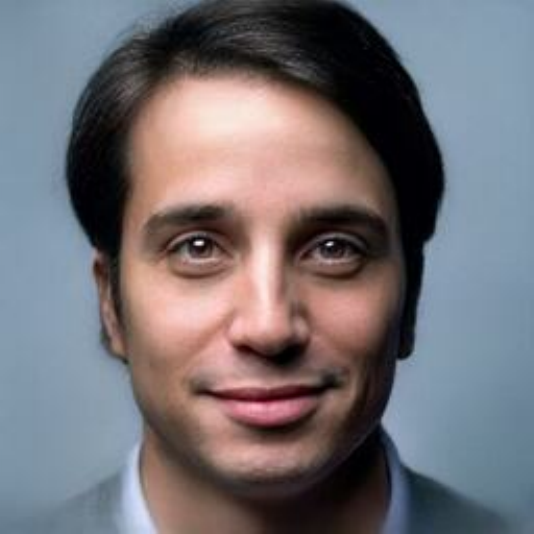} &
\includegraphics[width=0.17\columnwidth]{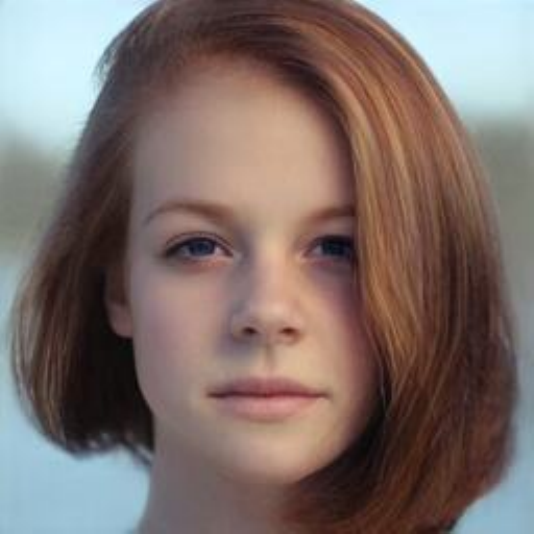} &
\includegraphics[width=0.17\columnwidth]{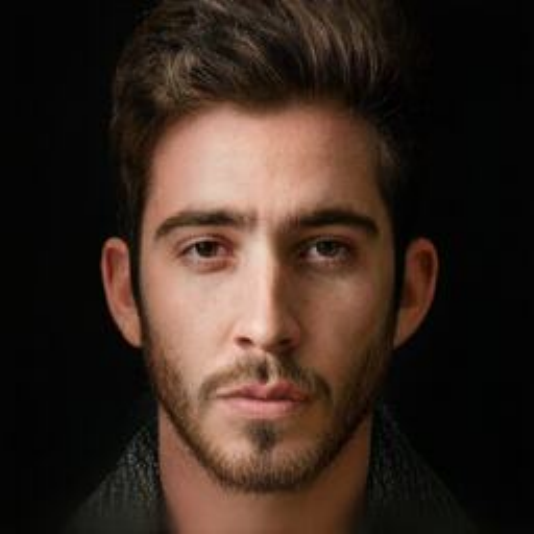} &
\includegraphics[width=0.17\columnwidth]{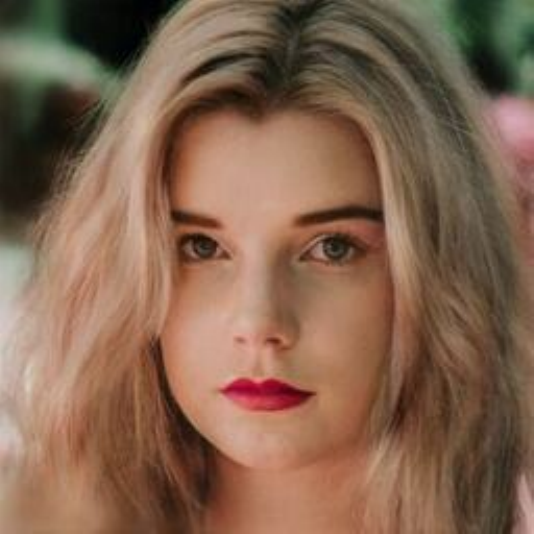} &
\includegraphics[width=0.17\columnwidth]{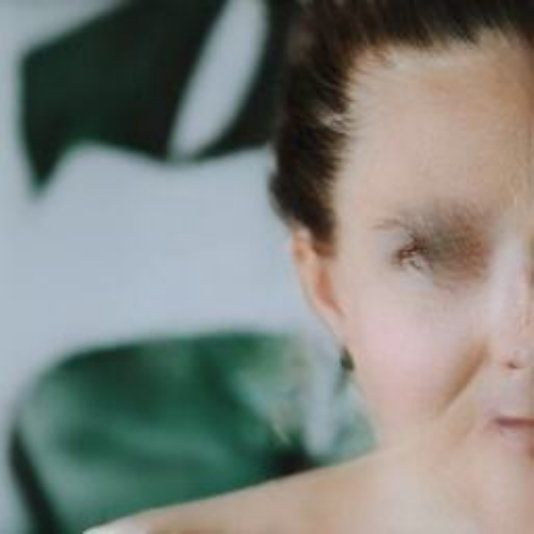} \\
\footnotesize LPIPS & \footnotesize $0.1249$ & \footnotesize $0.1083$ & \footnotesize $0.3031$  & \scriptsize $0.1249$ &  \footnotesize $0.0638$  \\
\footnotesize MSE & \footnotesize $0.0104$ & \footnotesize $0.0065$ & \footnotesize $0.0064$ & \footnotesize $0.0092$ &  \footnotesize $0.0053$  \\
\raisebox{1.2em}{\footnotesize \shortstack{PTI~\cite{roich2021pivotal} \\ \!w/\! $\ZPS$}} &
\includegraphics[width=0.17\columnwidth]{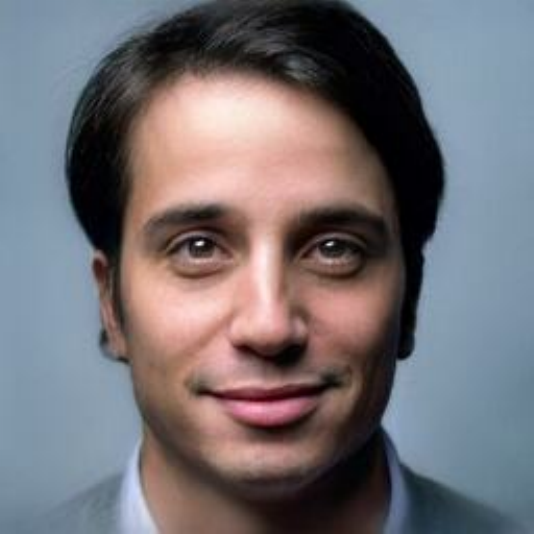} &
\includegraphics[width=0.17\columnwidth]{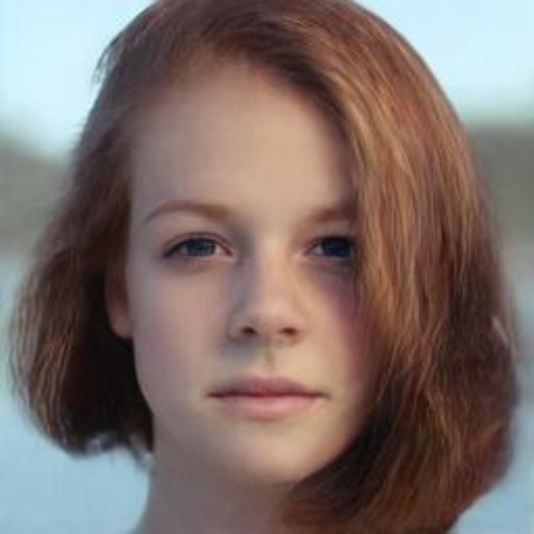} &
\includegraphics[width=0.17\columnwidth]{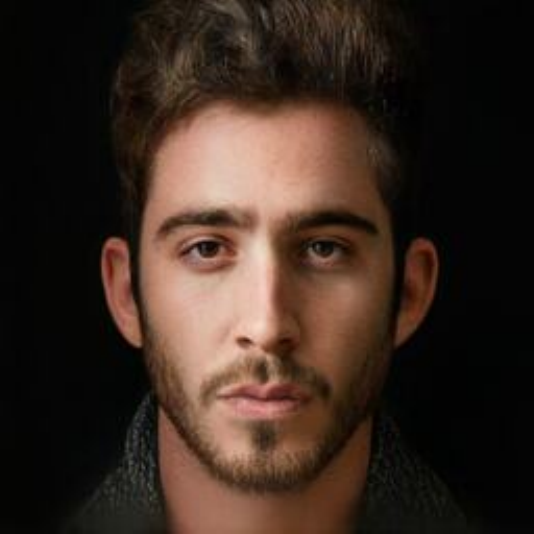} &
\includegraphics[width=0.17\columnwidth]{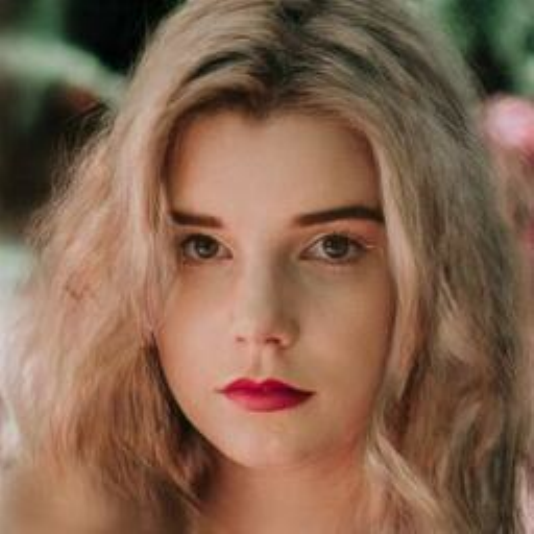} &
\includegraphics[width=0.17\columnwidth]{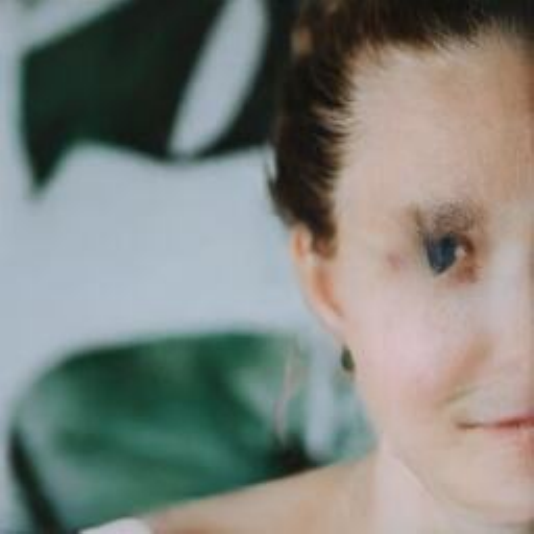} \\
\footnotesize LPIPS & \footnotesize $0.1095$ & \footnotesize $0.0852$ &  \footnotesize $0.3017$ & \footnotesize $0.1230$ & \footnotesize $0.0626$  \\
\footnotesize MSE & \footnotesize $0.0063$ & \footnotesize $0.0066$ & \footnotesize $0.0088$ & \footnotesize $0.0092$ & \footnotesize $0.0047$ \\
\raisebox{1.5em}{\footnotesize SAM~\cite{parmar2022spatially}} &
\includegraphics[width=0.17\columnwidth]{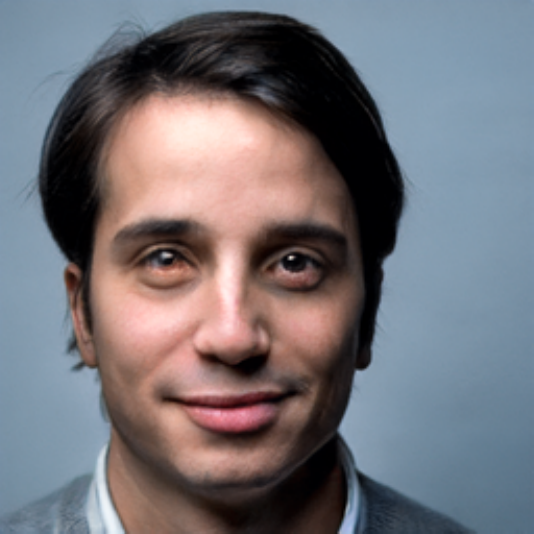} &
\includegraphics[width=0.17\columnwidth]{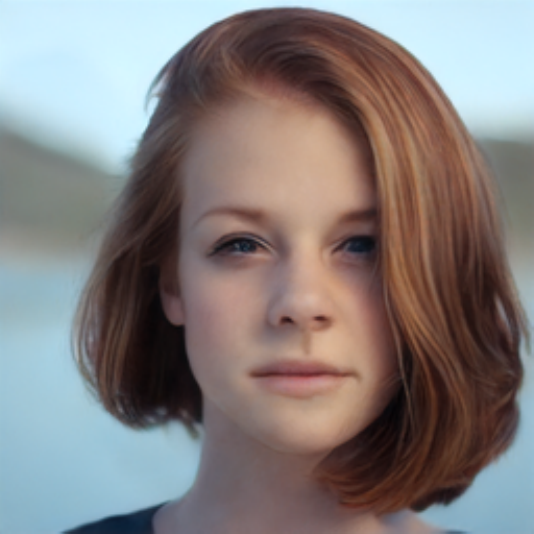} &
\includegraphics[width=0.17\columnwidth]{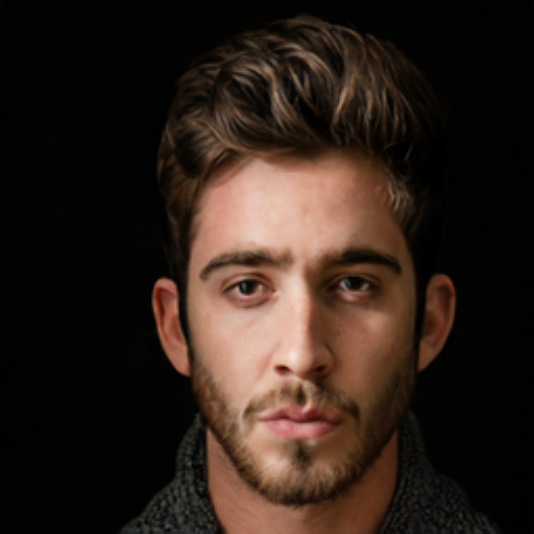} &
\includegraphics[width=0.17\columnwidth]{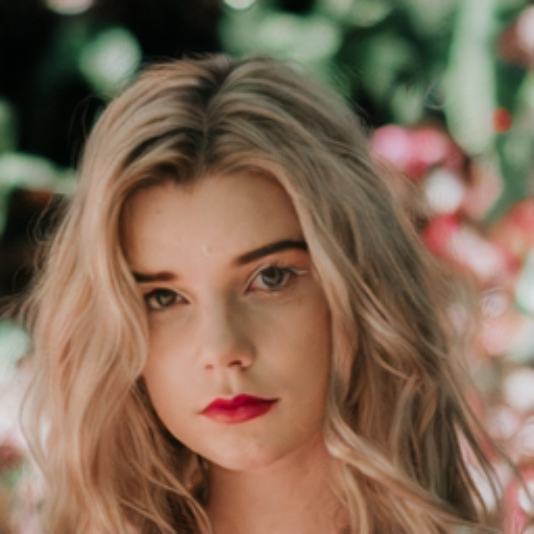} &
\includegraphics[width=0.17\columnwidth]{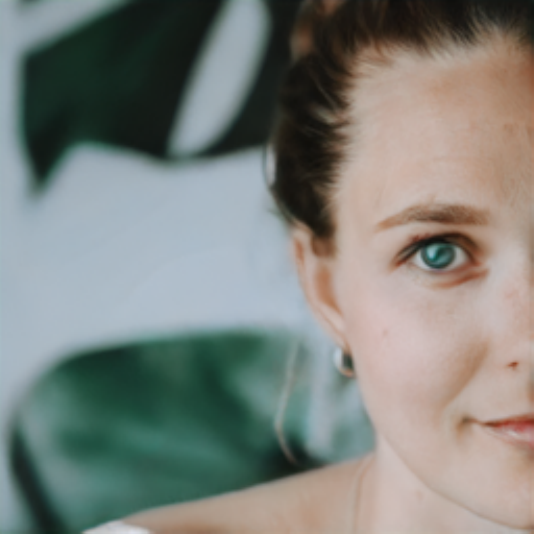} \\
\footnotesize LPIPS & \footnotesize $0.2111$ & \footnotesize $0.1169$ & \footnotesize $0.3802$ & \footnotesize $0.0809$ & \footnotesize $0.0609$ \\
\footnotesize MSE & \footnotesize $0.0098$ & \footnotesize $0.0057$ & \footnotesize $0.0074$ & \footnotesize $0.0037$ & \footnotesize $0.0023$ \\
\raisebox{1.2em}{\footnotesize \shortstack{SAM~\cite{parmar2022spatially} \\ \!w/\! $\ZPS$}}  &
\includegraphics[width=0.17\columnwidth]{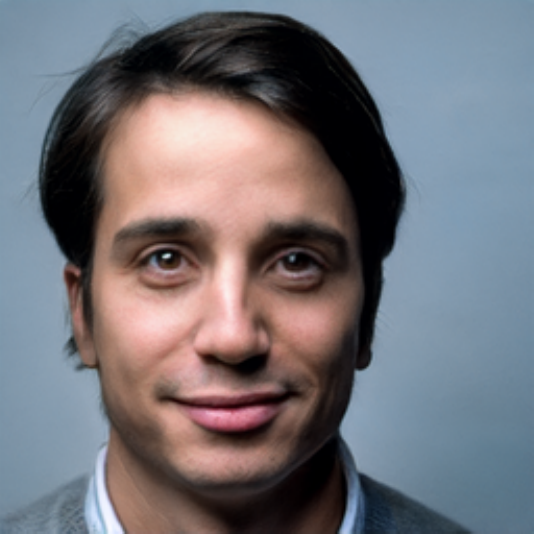} &
\includegraphics[width=0.17\columnwidth]{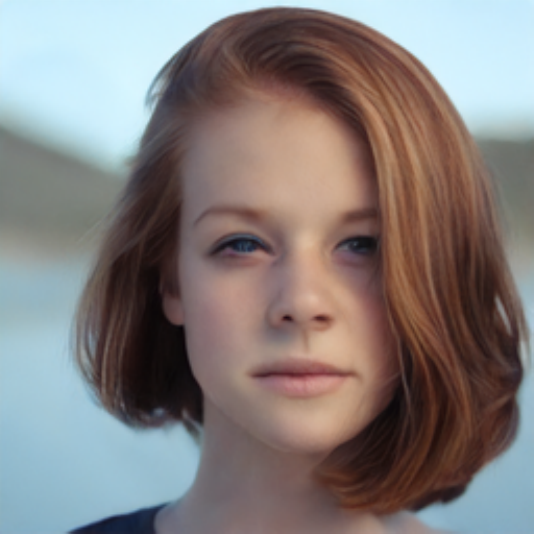} &
\includegraphics[width=0.17\columnwidth]{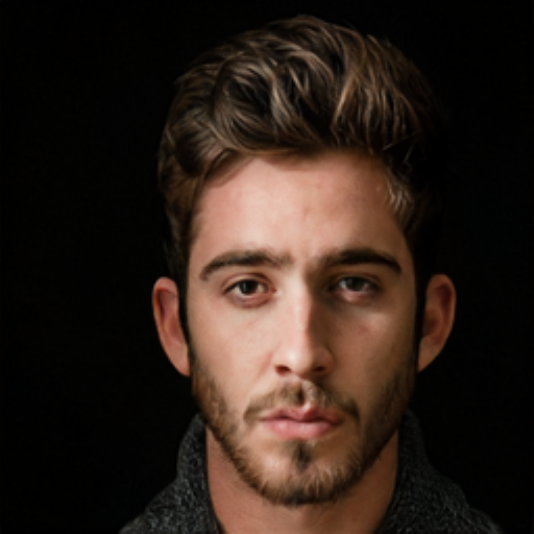} &
\includegraphics[width=0.17\columnwidth]{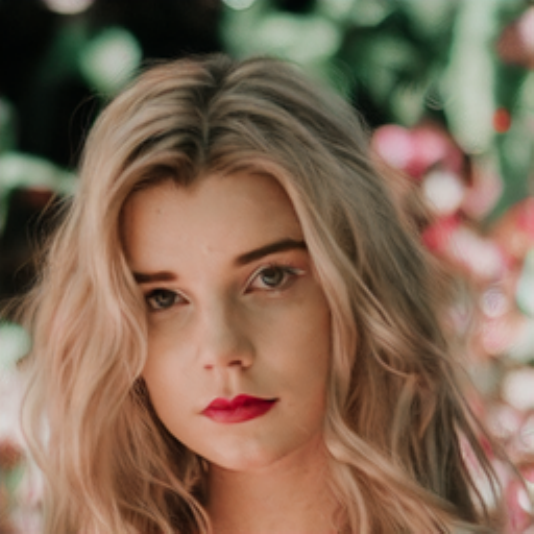} &
\includegraphics[width=0.17\columnwidth]{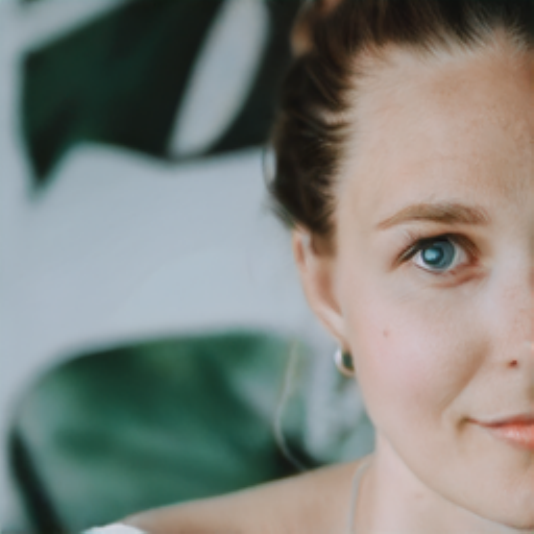} \\
\footnotesize LPIPS &  \footnotesize $0.1987$ & \footnotesize $0.1247$  & \footnotesize $0.3838$ & \footnotesize $0.0793$ & \footnotesize $0.0584$ \\
\footnotesize MSE & \footnotesize $0.0095$ & \footnotesize $0.0060$ & \footnotesize $0.0080$ & \footnotesize $0.0041$ &  \footnotesize $0.0025$ \\
    \end{tabular}\egroup
\caption{Reconstruction comparisons on state-of-the-arts. The 1st and 3rd rows are inverted results of PTI and SAM. The 2nd and 4th rows are inverted results of the methods that use $\ZPS$ space instead of $\WS$ or $\WPS$. It indicates we can replace original latent space to $\ZPS$ without loosing reconstruction quality to improve editing quality.}\label{fig:sota}
\end{figure}

\begin{figure}[t]
  \centering
    \bgroup 
    \def\arraystretch{0.2} 
    \setlength\tabcolsep{2pt}
    \begin{tabular}{cccccccc}
inversion & eyeglass & smile & invert & eyeglass & smile \\
\includegraphics[width=0.15\columnwidth]{rebuttal/sam_z/sample1.pdf} &
\includegraphics[width=0.15\columnwidth]{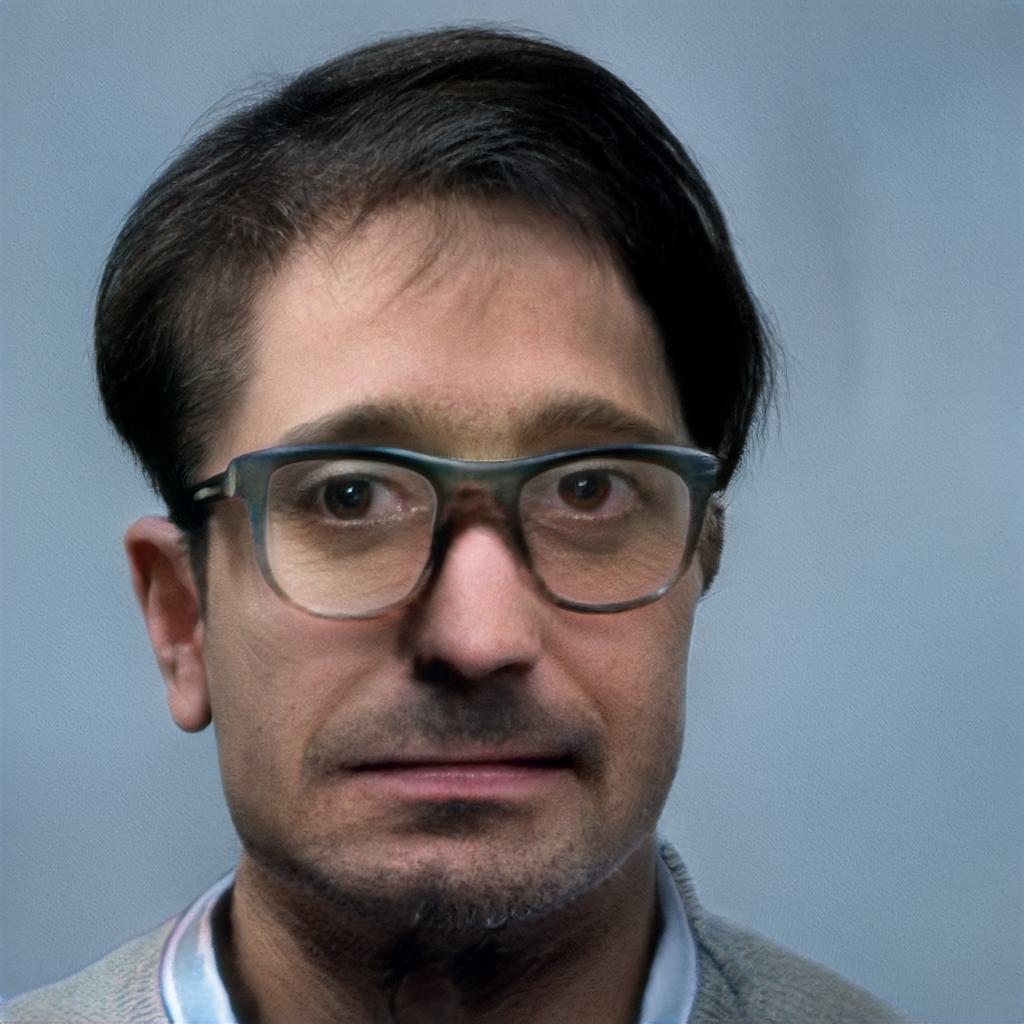} &
\includegraphics[width=0.15\columnwidth]{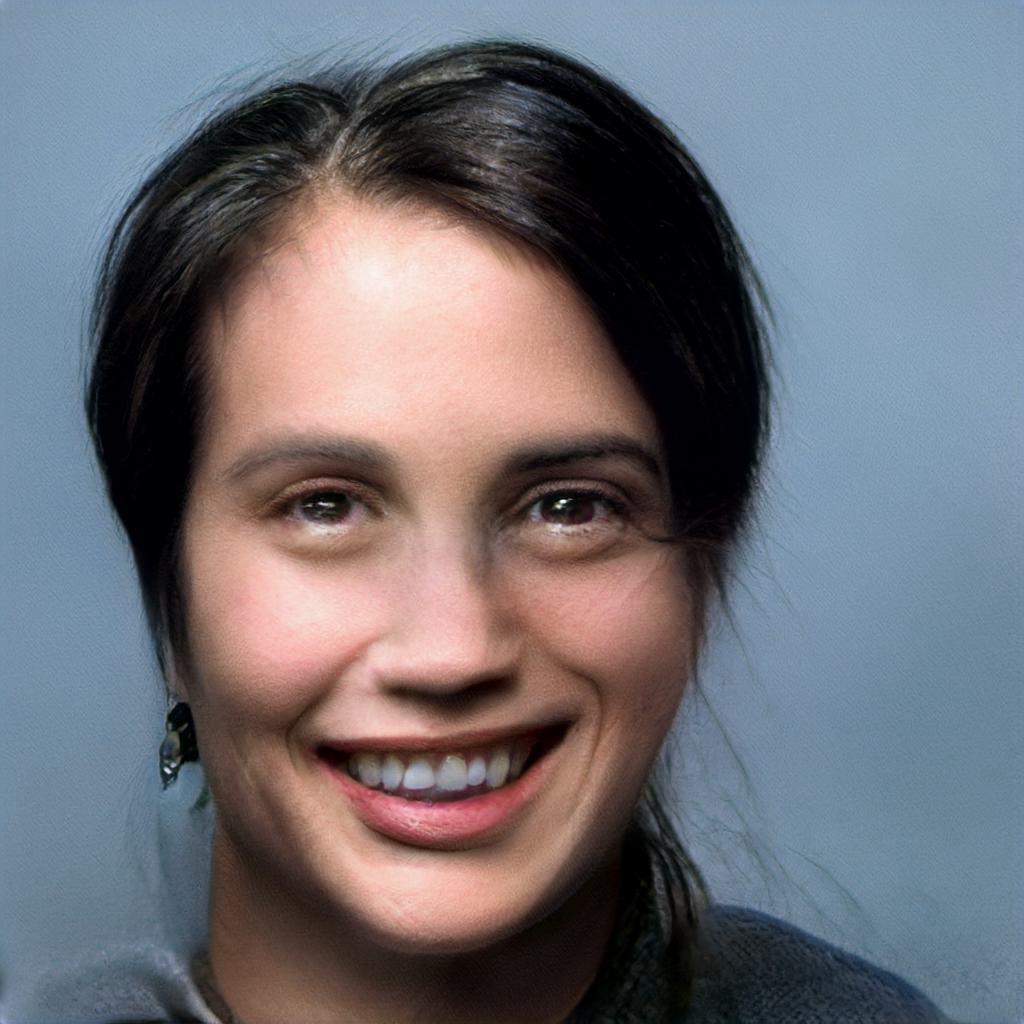} &
\includegraphics[width=0.15\columnwidth]{rebuttal/sam_w/sample1.pdf} &
\includegraphics[width=0.15\columnwidth]{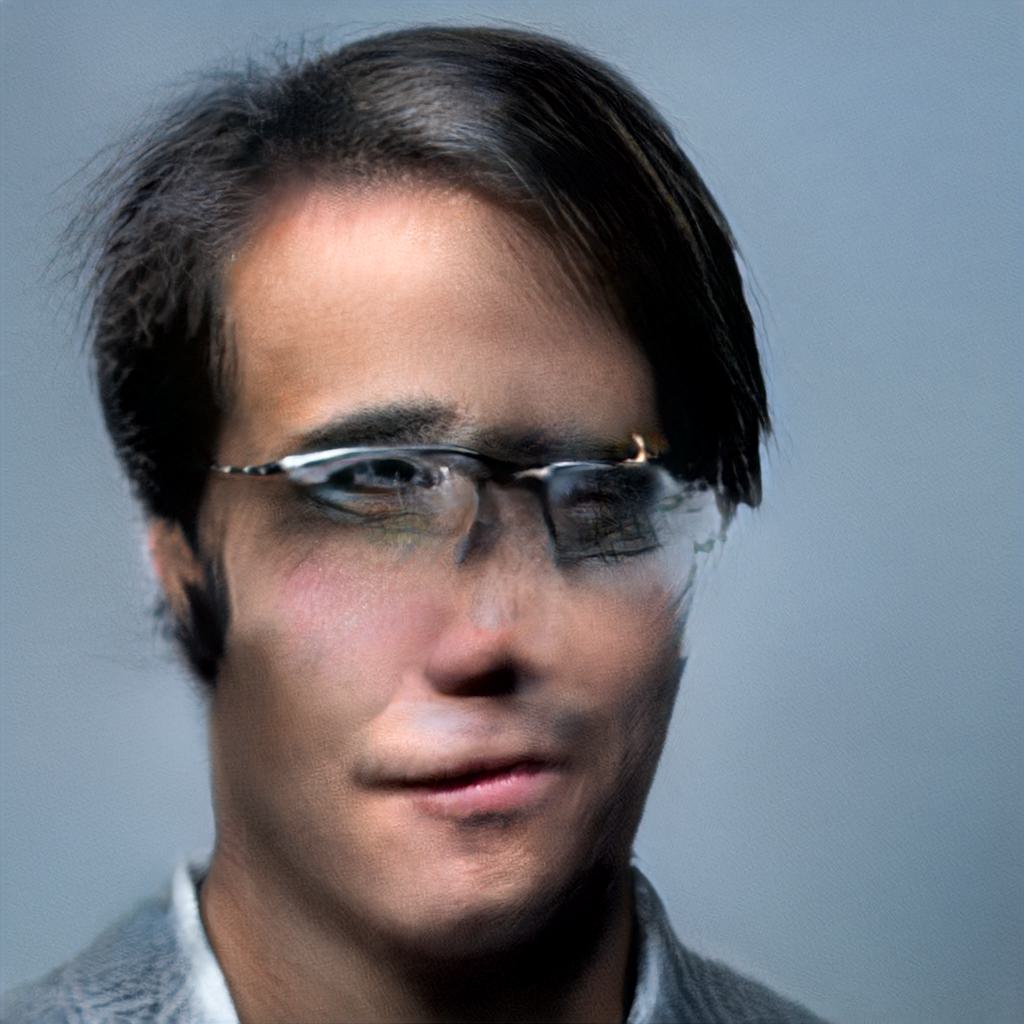} &
\includegraphics[width=0.15\columnwidth]{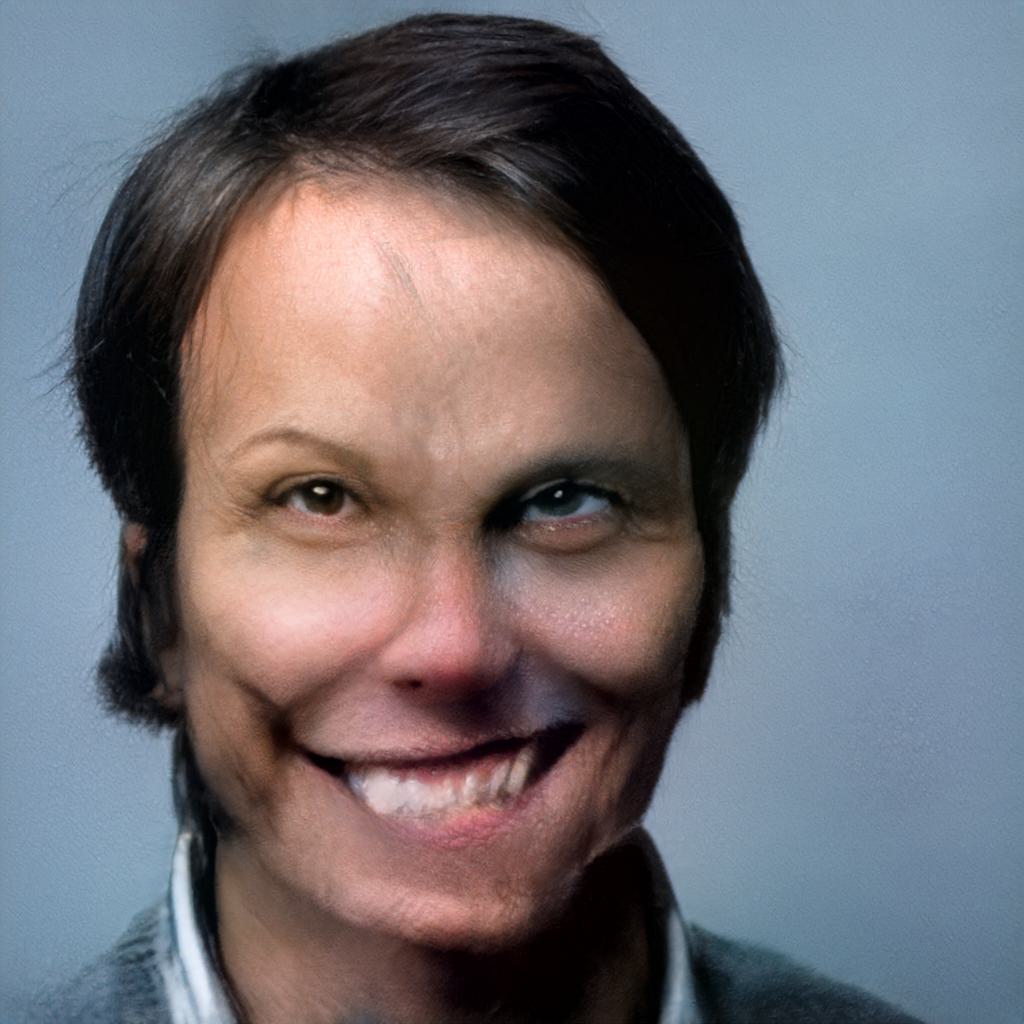} \\\\
\multicolumn{3}{c}{SAM~\cite{parmar2022spatially} w/ $\ZPS$} & \multicolumn{3}{c}{SAM~\cite{parmar2022spatially}} \\\\
inversion & age & smile & invert & age & smile \\
\includegraphics[width=0.15\columnwidth]{rebuttal/pti_z/sample5.pdf} &
\includegraphics[width=0.15\columnwidth]{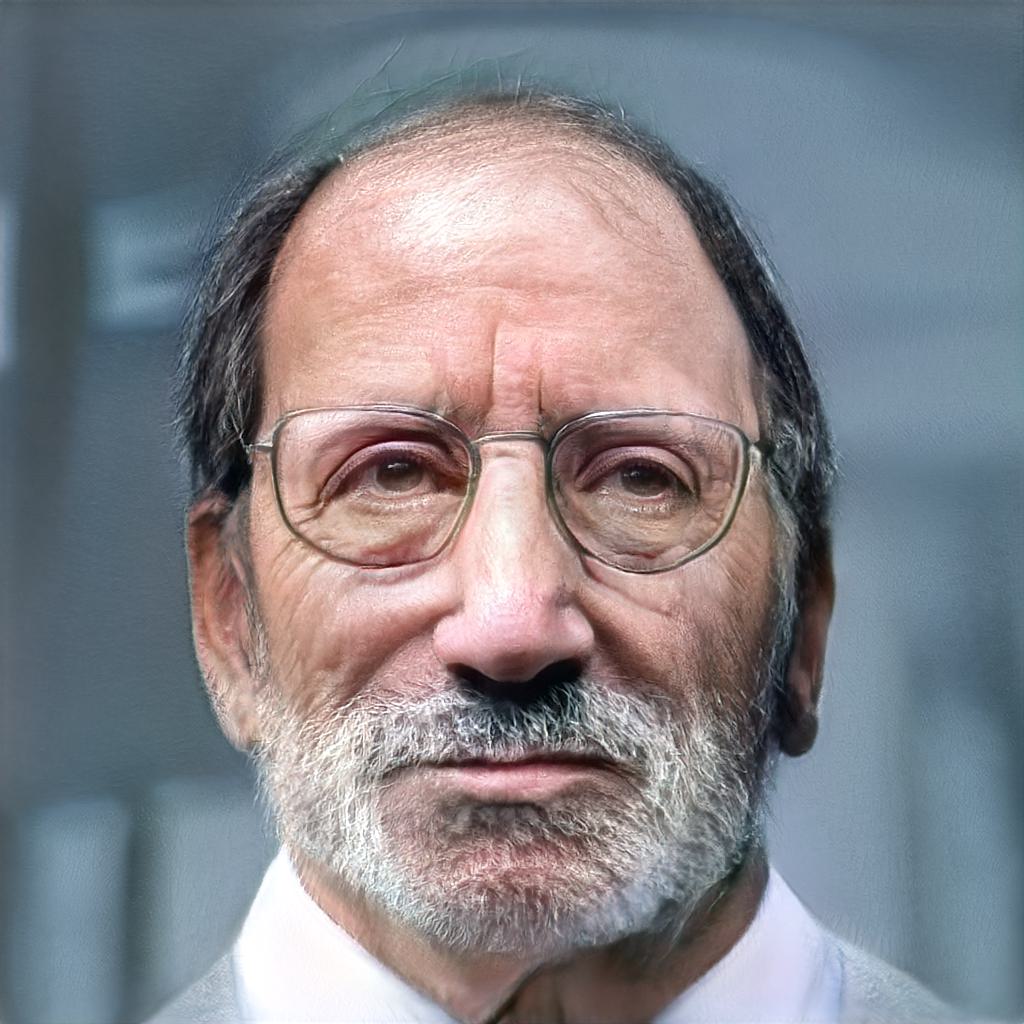} &
\includegraphics[width=0.15\columnwidth]{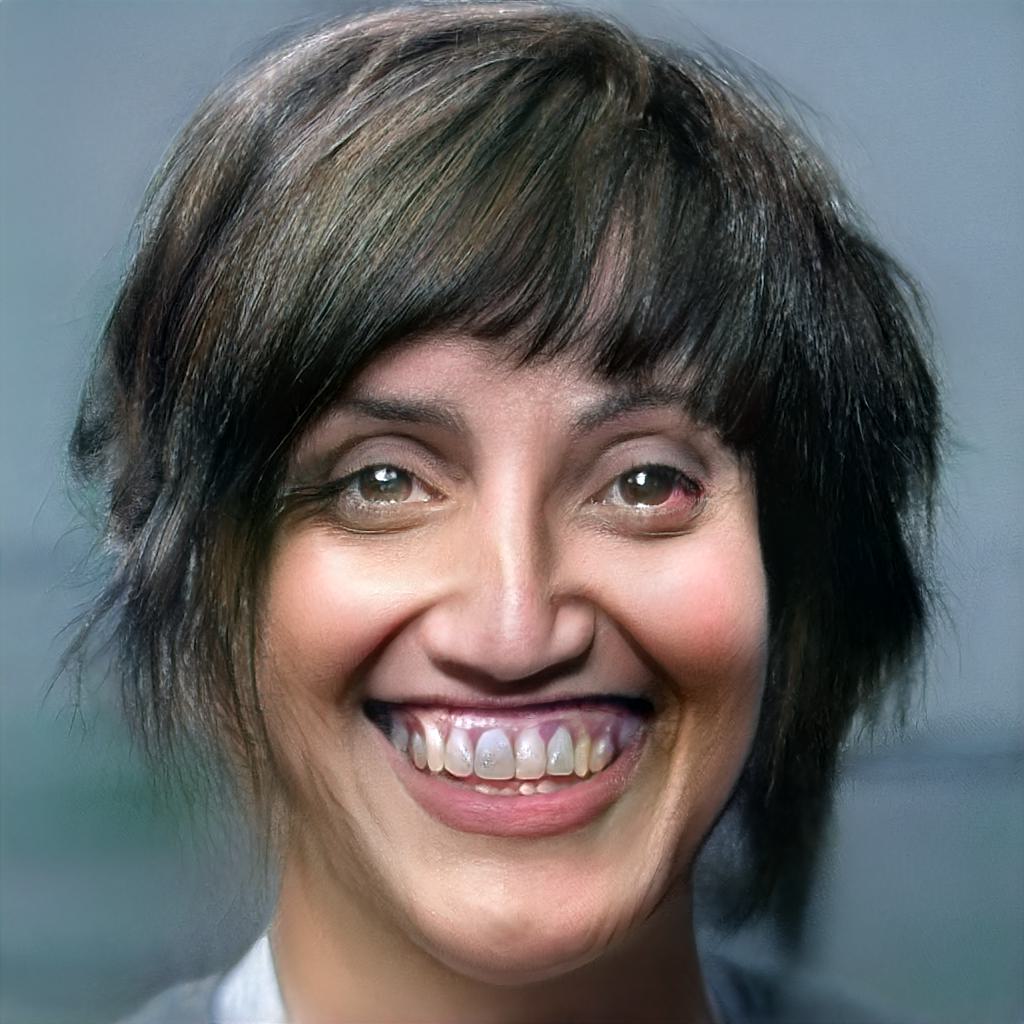} &
\includegraphics[width=0.15\columnwidth]{rebuttal/pti_w/sample5.pdf} &
\includegraphics[width=0.15\columnwidth]{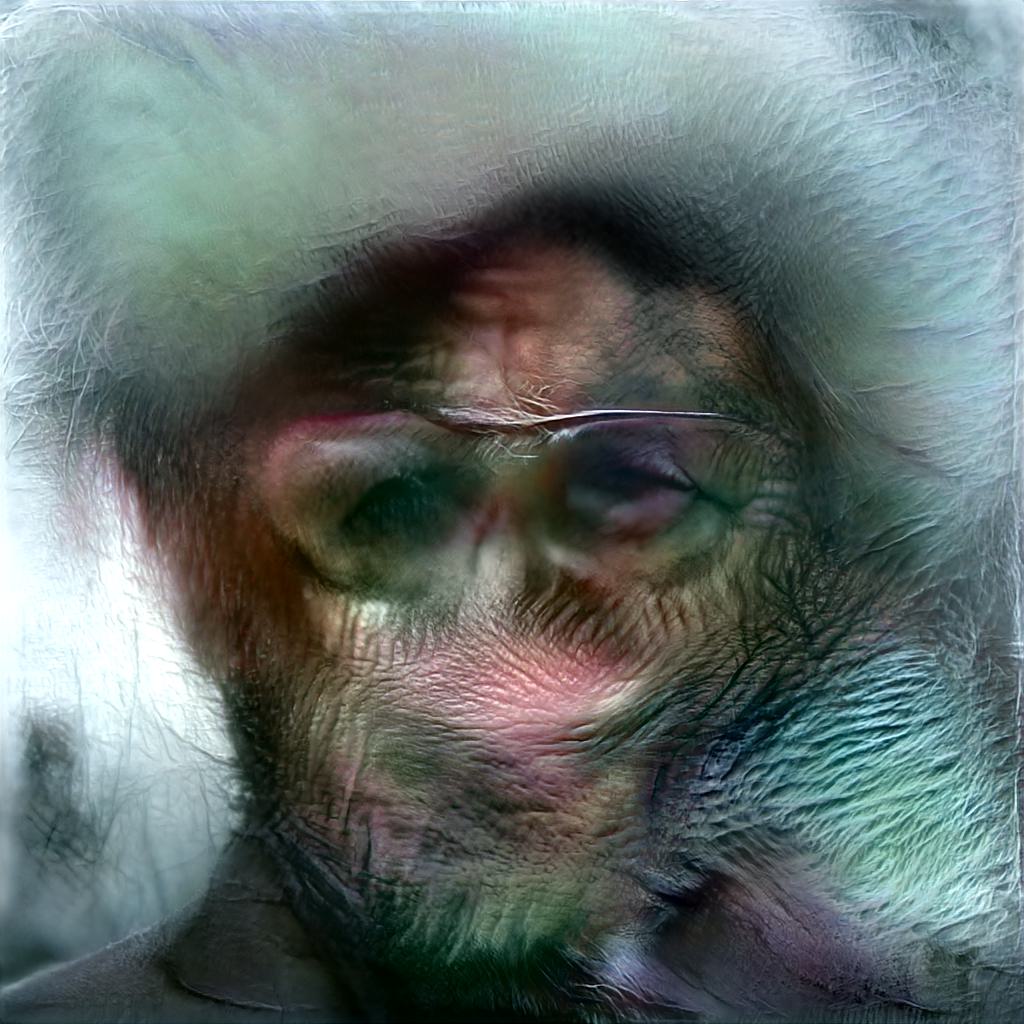} &
\includegraphics[width=0.15\columnwidth]{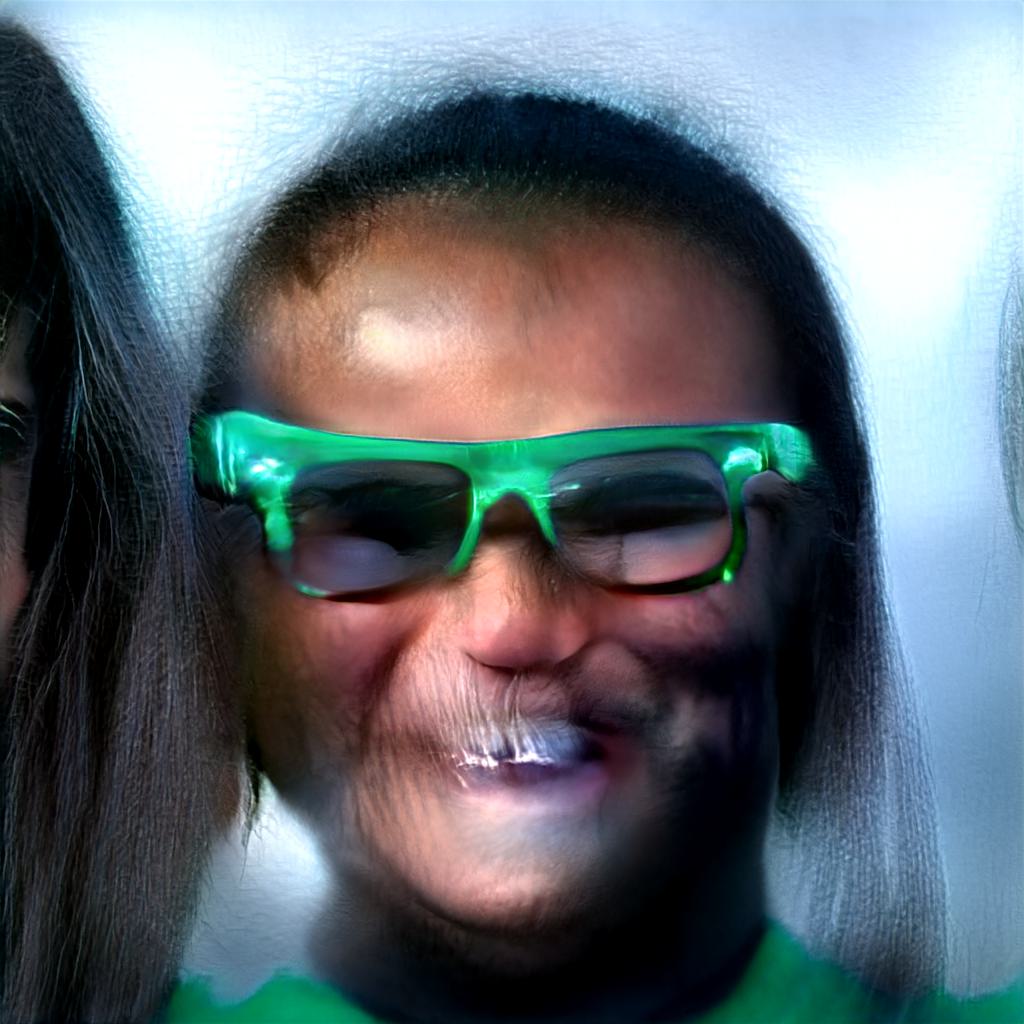} \\
\multicolumn{3}{c}{PTI~\cite{roich2021pivotal} w/ $\ZPS$} & \multicolumn{3}{c}{PTI~\cite{roich2021pivotal}}
    \end{tabular}\egroup
\caption{Editing comparison on SAM and PTI. By replacing $\WPS$ in SAM or $\WS$ in PTI to $\ZPS$ avoid harming perceptual quality of edited images. }\label{fig:editing_sota}
\end{figure}

\begin{figure}[h]
  \centering
    \bgroup 
    \def\arraystretch{0.2} 
    \setlength\tabcolsep{1.8pt}
    \begin{tabular}{ccccc}
& inversion & +big noise & +age & +smile \\
\raisebox{1.4em}{\shortstack{SAM~\cite{parmar2022spatially} \\ w/ $\ZPS$}} &
\includegraphics[width=0.17\columnwidth]{rebuttal/sam_z/sample1.pdf} 
& \includegraphics[width=0.17\columnwidth]{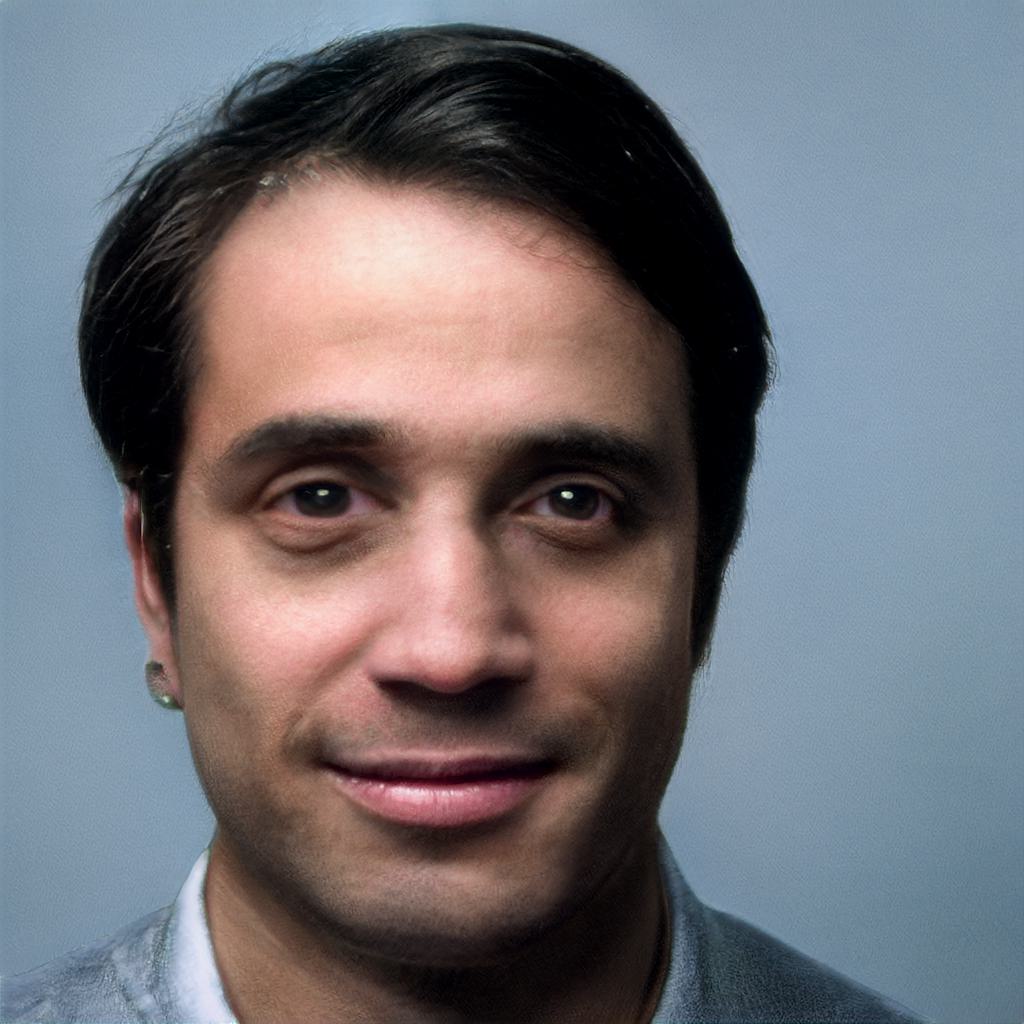} &
\includegraphics[width=0.17\columnwidth]{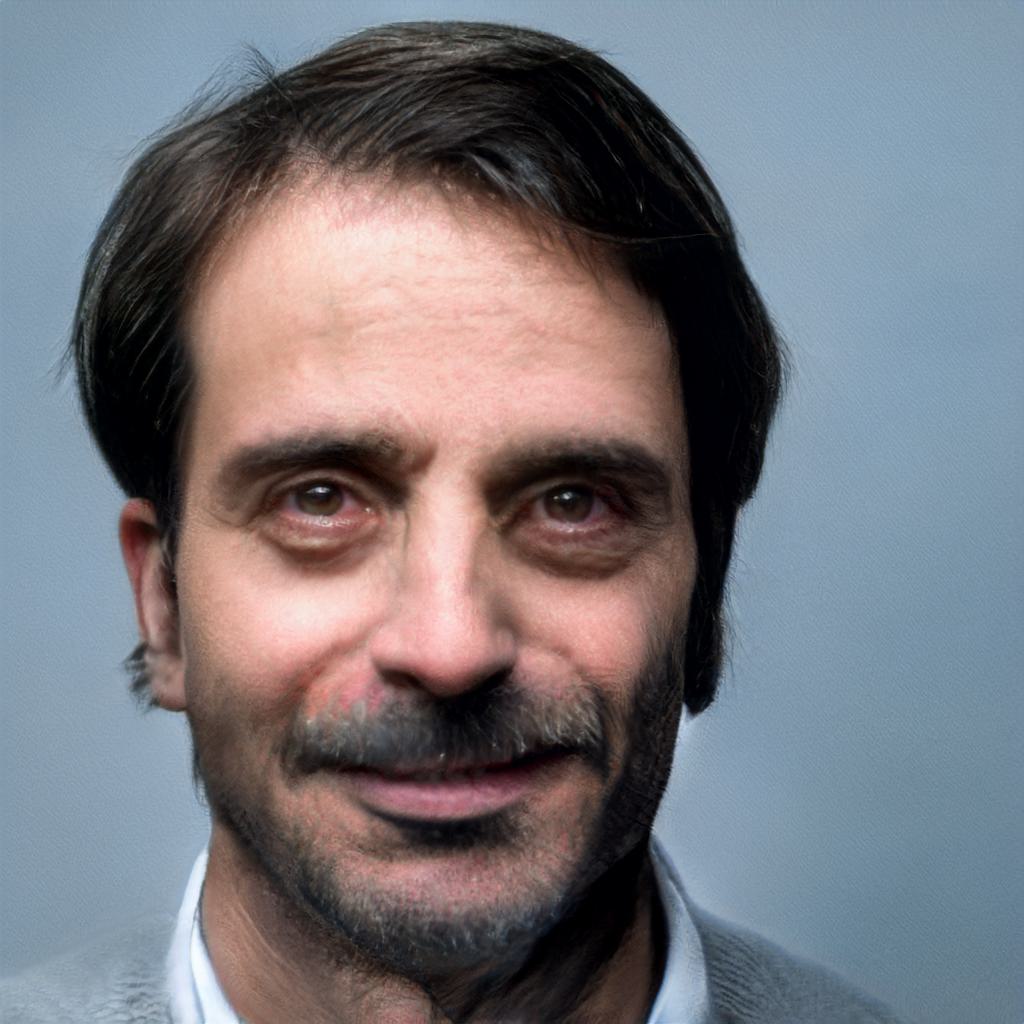} &
\includegraphics[width=0.17\columnwidth]{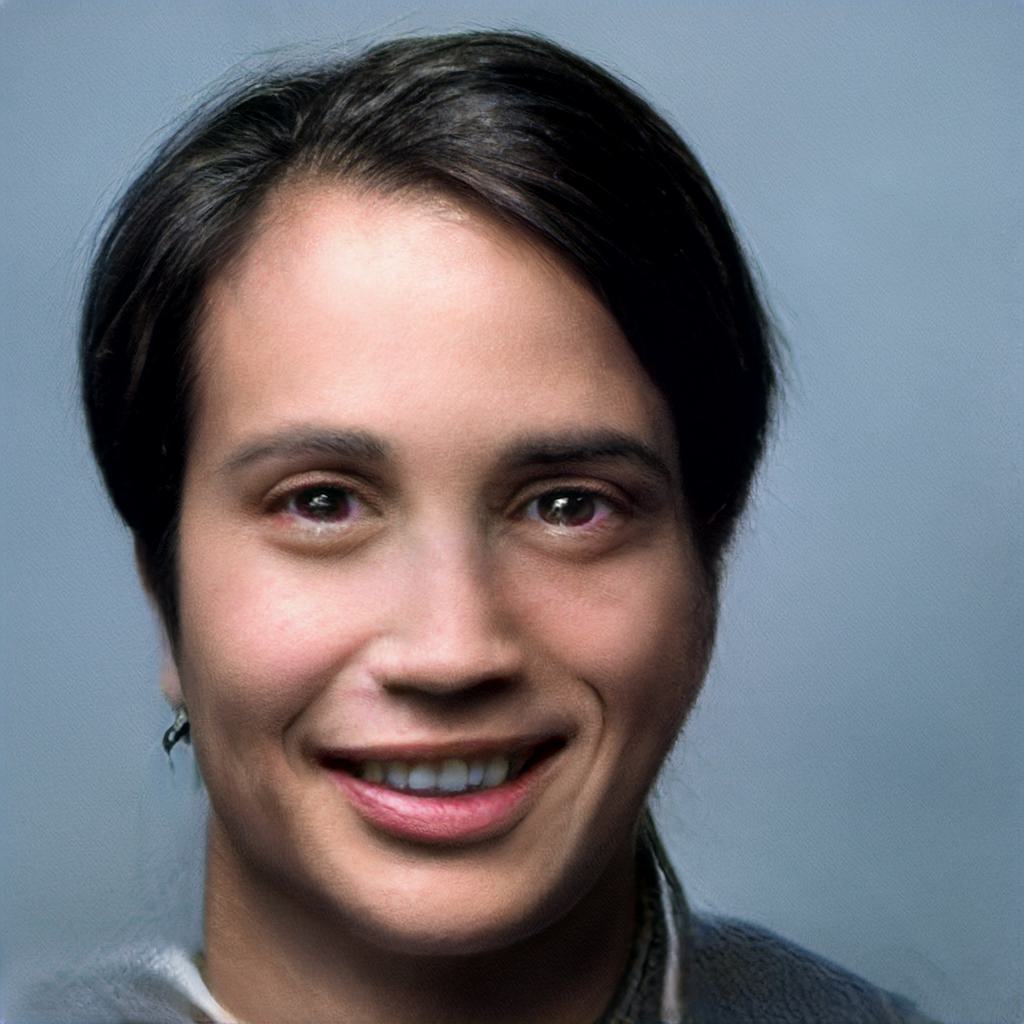} \\
 & inversion & +smile & pale\,skin & -age \\
\multirow{3}{*}{$\FZS$} & \includegraphics[width=0.17\columnwidth]{interfacegan/zp_sample1_invert.pdf} & 
\includegraphics[width=0.17\columnwidth]{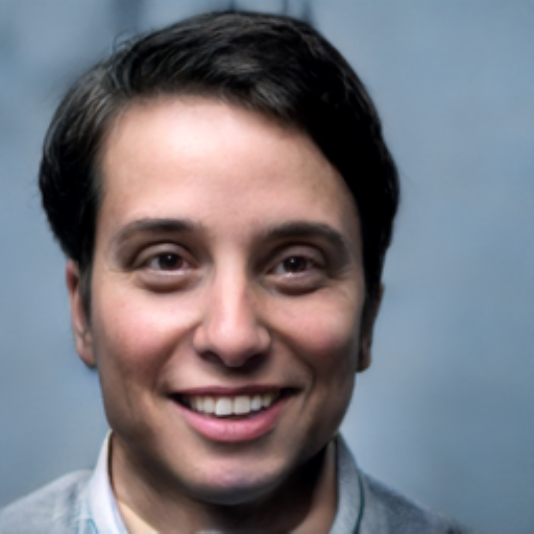} &
\includegraphics[width=0.17\columnwidth]{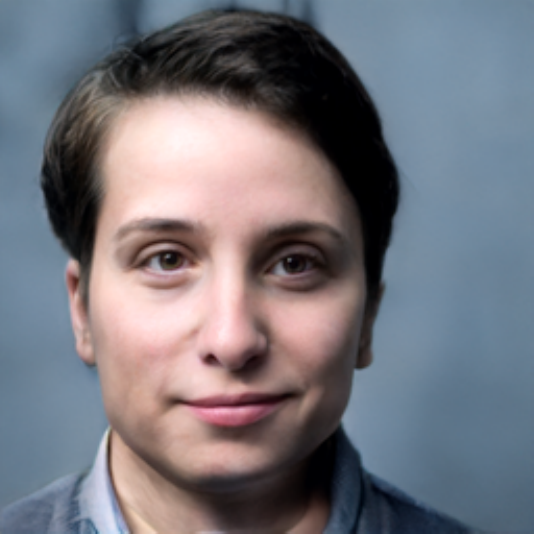} &
\includegraphics[width=0.17\columnwidth]{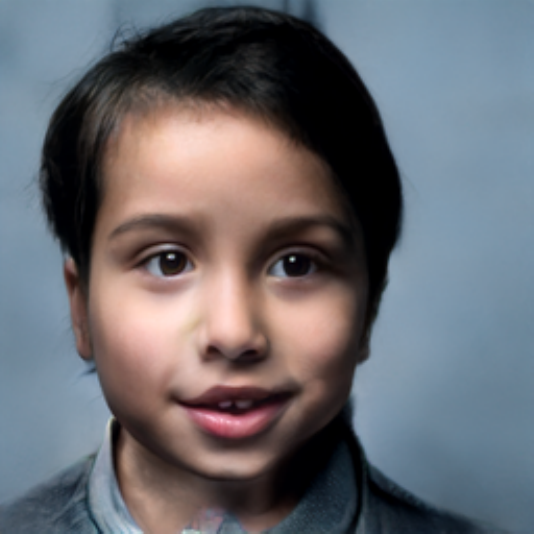} \\
&& blond\,hair & narrow\,eye & mustashe \\
&&\includegraphics[width=0.17\columnwidth]{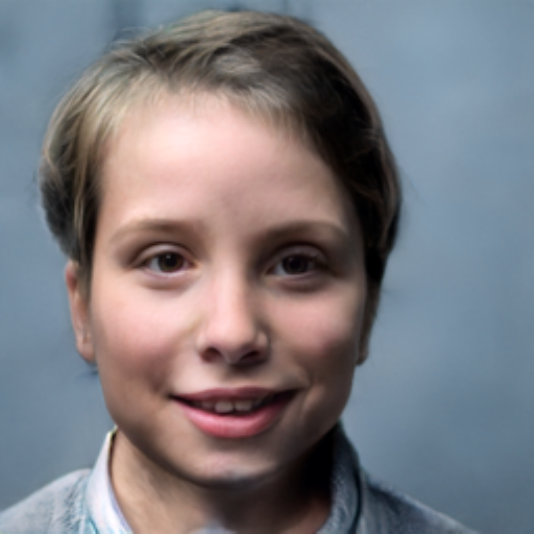} &
\includegraphics[width=0.17\columnwidth]{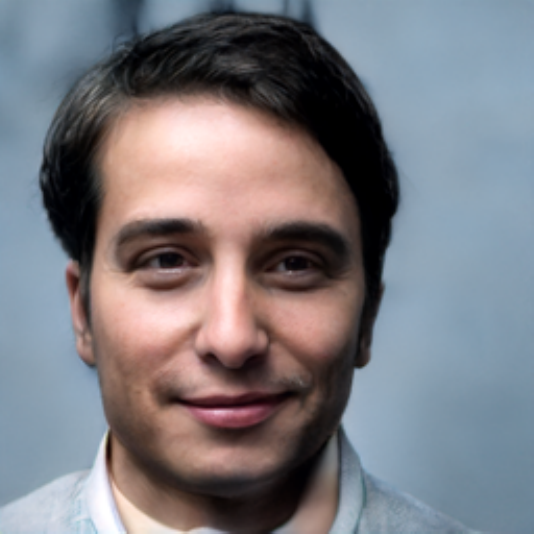} &
\includegraphics[width=0.17\columnwidth]{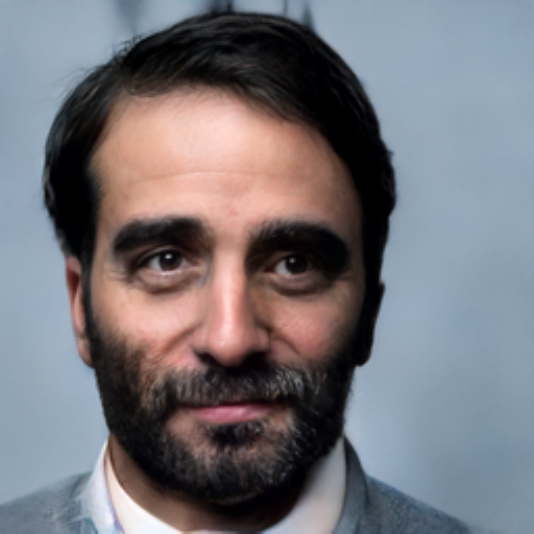} \\
    \end{tabular}\egroup
\caption{Editing examples with InterfaceGAN directions on SAM + $\ZS$ and $\FZS$.}\label{fig:editing_example}
\end{figure}

\noindent 
\textbf{Integration $\FZS$ into state-of-the-art GAN inversion}.
We further demonstrate the effectiveness of our $\ZPS$ space. \Cref{fig:sota} shows reconstructed images by PTI~\cite{roich2021pivotal}, SAM~\cite{parmar2022spatially}, and $\ZPS$ version of them. We can see that the use of $\ZPS$ on PTI and SAM does not sacrifice reconstruction performance.
\Cref{fig:editing_sota} shows that integrating $\ZPS$ space into SoTA inversion methods relaxes editing distortions.

We additionally provide editing examples on SAM with $\ZPS$ and $\FZS$ with InterfaceGAN directions in \cref{fig:editing_example}.
The figure shows that the $\ZPS$ space naturally performs latent editing.

 \section{Conclusion}
 \label{sec:conclusion}

We revisit $\ZS$ space for GAN inversion to yield a better trade-off between reconstruction quality and editing quality. We integrate bounded latent space $\ZPS$ with the hyperspherical prior instead of $\WPS$ into the space with rich representative capacity, resulting in the presented space (\eg, $\FZS$). Our thorough experiments on PTI, SAM, $\FWS$, and $\FSS$ demonstrate that we can preserve perceptual quality of edited images while maintaining sufficient reconstruction quality on par with baseline methods by replacing unbounded space (\eg, $\WPS$) to $\ZPS$.

\noindent \textbf{Broader impact.}
In this study, we investigate the hyperspherical prior of unconditional 2D StyleGAN. 
While diffusion models have undergone rapid development, StyleGAN-based have demonstrated competitiveness in 3D generation~\cite{Chan2022,3dgp} and text-to-image generation~\cite{sauer2023stylegan}. Besides unconditional 2D StyleGAN, this study has the potential to contribute to these approaches.
Additionally, our approach offers a promising way to enhance inversion techniques for diffusion models with a hyperspherical prior such as the CLIP~\cite{radford2021learning} embedding since such prior also has a bounded nature like $\ZPS$.

\noindent \textbf{Limitation.} 
Our approach achieves a better trade-off between reconstruction quality and editing quality than popularly used latent spaces. However, since most attempts for GAN inversion and latent editing have focused on the $\WS$, $\WPS$, and $\SSp$ spaces, the semantic directions in $\ZS$ and $\ZPS$ are less disentangled than in the well-explored spaces.

{\small
\bibliographystyle{ieee_fullname}
\bibliography{egbib}
}

\end{document}